\documentclass{article}

\usepackage{microtype}
\usepackage{graphicx}
\usepackage{subcaption}
\usepackage{booktabs} 

\usepackage{hyperref}

\usepackage[preprint]{icml2026}

\usepackage{amsmath}
\usepackage{amssymb}
\usepackage{mathtools}
\usepackage{amsthm}
\usepackage{times,latexsym}
\usepackage{url}
\usepackage{xurl} 
\usepackage[T1]{fontenc}
\usepackage{graphicx}
\usepackage{booktabs}
\usepackage{xspace}

\usepackage{mfirstuc,tabulary}

\usepackage[capitalize,noabbrev]{cleveref}

\theoremstyle{plain}

\theoremstyle{definition}

\theoremstyle{remark}

\usepackage[textsize=tiny]{todonotes}

\newcommand{\model}{\textsc{SimAI}\xspace}

\icmltitlerunning{SimAI: Mechanistic User Simulation through Psychological Network Models}

\begin{document}

\twocolumn[
  \icmltitle{SimAI: Mechanistic User Simulation through Psychological Network Models}




    \icmlsetsymbol{equaladvising}{$\dagger$}
    
    \begin{icmlauthorlist}
      \icmlauthor{Cheng Li}{cse}
      \icmlauthor{Adhyyan Narang}{ece}
      \icmlauthor{Jiayi Yuan}{cse}
      \icmlauthor{John F. Buss}{med}
      \icmlauthor{Jina Suh}{cse}
      \icmlauthor{Devon Sandel-Fernandez}{med}
      \icmlauthor{Yulia Tsvetkov}{cse}
      \icmlauthor{Adam M. Kuczynski}{med,equaladvising}
      \icmlauthor{Tim Althoff}{cse,equaladvising}
    \end{icmlauthorlist}
    
    \icmlaffiliation{cse}{
      Paul G. Allen School of Computer Science \& Engineering,
      University of Washington, Seattle, WA, USA
    }
    \icmlaffiliation{ece}{
      Department of Electrical \& Computer Engineering,
      University of Washington, Seattle, WA, USA
    }
    \icmlaffiliation{med}{
      Department of Psychiatry and Behavioral Sciences,
      University of Washington School of Medicine,
      Seattle, WA, USA
    }
    
    \icmlcorrespondingauthor{Cheng Li}{chengli3@cs.washington.edu}
    
    \icmlkeywords{Machine Learning, User Simulation}
    
    \vskip 0.3in

]



\printAffiliationsAndNotice{
 \textsuperscript{$\dagger$}Co-senior authors,
}  

\begin{abstract}
Realistic user simulation enables the study of human interactions when direct experimentation is costly or potentially harmful. However, existing simulators largely overlook the psychological mechanisms that drive user behavior. 
We introduce \model, a framework for \emph{mechanistic user
simulation} that models how psychological factors interact
to shape user responses.
To overcome the challenge of supervising open-ended behavior,
we propose a graph reconstruction objective that provides
graph-level rewards based on how well simulated conversations
reflect the underlying psychological relationships.
To support this approach, we construct \textsc{PSYCHE}, a dataset of $18,576$ profile--network pairs derived from $516$ psychotherapy cases.
Automatic evaluations show that \model achieves greater simulation and behavioral diversity than existing patient simulators while maintaining strong profile alignment. 
In an interactive study with $30$ mental-health professionals
and trainees, \model receives the highest average ratings for conversational authenticity, realistic help-seeking behavior, and clinical training utility.
Beyond simulation quality, \model enables realistic clinical
training and uncovers safety failures in prolonged AI
interactions with vulnerable users.
Code is available at \url{https://github.com/behavioral-data/ANGEL-UserSim}.\footnote{The dataset and trained models will be released upon completion of
licensing with the source journal.}
\end{abstract}

\begin{figure*}[t]
    \centering
    \includegraphics[width=\textwidth]{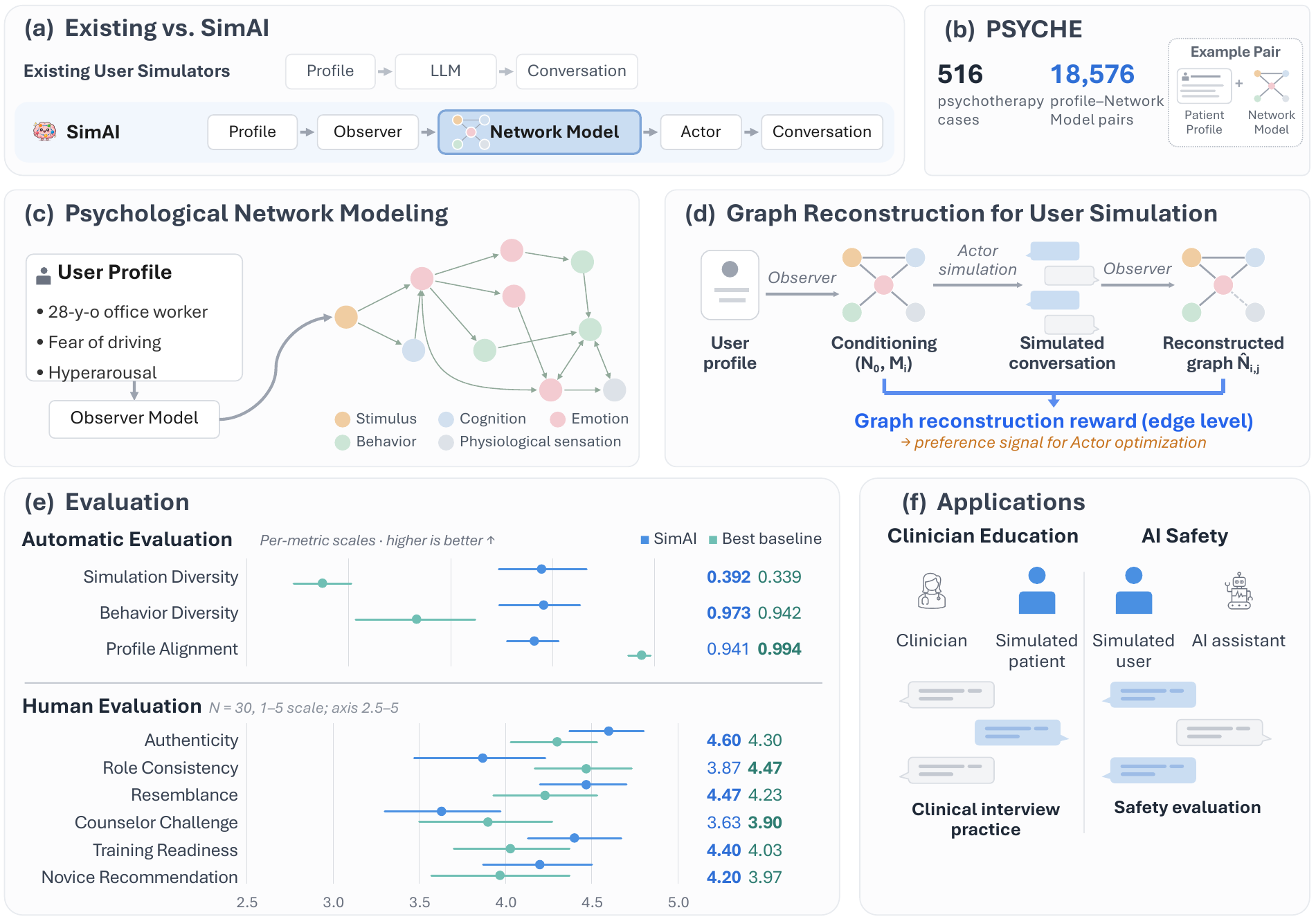}
    \caption{
    \textbf{Overview of \model.}
    \textbf{(a)} Unlike existing user simulators that generate conversations
    directly from profiles, \model represents underlying psychological
    mechanisms through a network model.
    \textbf{(b)} \textsc{PSYCHE} contains 18,576 paired profile--network
    model samples derived from 516 psychotherapy cases.
    \textbf{(c)} The Observer constructs a psychological network model
    from a user profile; the example is simplified for clarity.
    \textbf{(d)} Selected graph relationships are masked to condition
    the Actor, while graph reconstruction provides training feedback.
    \textbf{(e)} Automatic and human evaluations compare \model
    with the best baseline for each metric. Points indicate means
    and error bars show 95\% confidence intervals.
    \textbf{(f)} Applications in clinician education and AI safety.
    }
    \label{fig:overview}
\end{figure*}

\begin{figure*}[t]
    \centering
    \includegraphics[width=.95\textwidth]{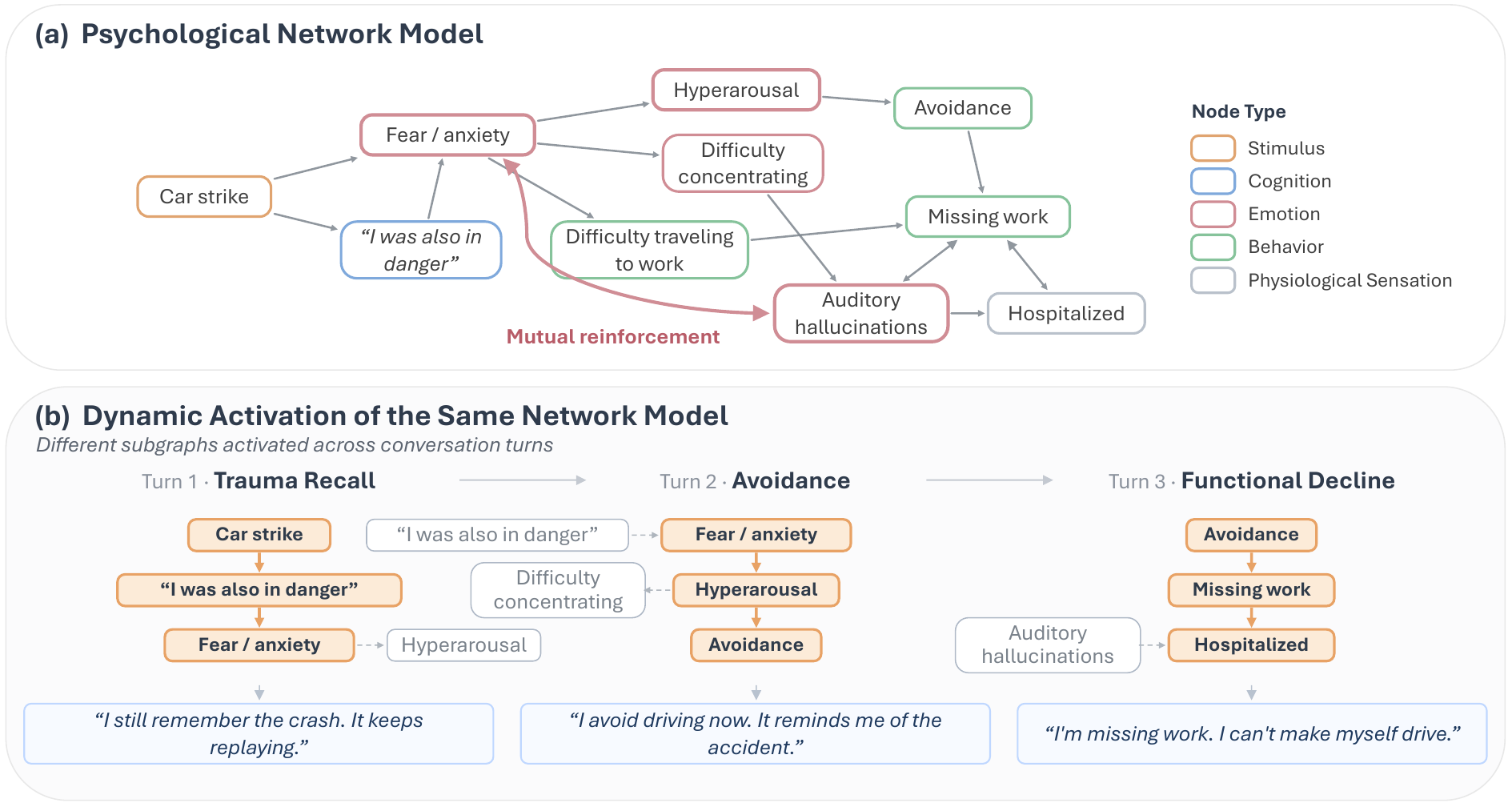}
    \caption{
    \textbf{Network model and its activation during conversation.}
    \textbf{(a)} Nodes represent stimuli, cognitions, emotions, behaviors, and
    physiological sensations; edges represent causal, triggering, or reinforcing
    relationships.
    \textbf{(b)} Different conversational turns activate different subgraphs of
    the underlying network model to ground the Actor's responses.
    }
    \label{fig:network_model}
\end{figure*}

\section{Introduction}


Understanding and anticipating human behavior is important for both developing AI systems that better serve users and preparing professionals for complex human interactions~\cite{he2023survey,kononowicz2019virtual,jiang2026genai}. 
However, studying these interactions with real people is costly and difficult to scale, given the diversity of individual backgrounds, needs, and behaviors~\cite{hu2026simbench,chopra2026beyond}.
These challenges are especially pronounced in high-stakes settings involving vulnerable populations, such as individuals experiencing mental health difficulties, where privacy concerns, ethical constraints, and potential harm further limit direct
experimentation~\cite{rahsepar2025exploring,diel2026scoping}.
Realistic user simulation offers a promising alternative, enabling controlled exploration of diverse user behaviors and conversational scenarios.
Such simulations can support professional training, such as clinicians interviewing simulated patients, as well as training and evaluating foundation models across diverse users~\cite{wang2024patient,dou2025simulatorarena}.

Existing user simulators have made progress in generating plausible user behaviors by conditioning LLMs on user profiles and behavioral characteristics~\cite{chuang2025debate,lu2026can,kolluri2025finetuning,binz2025foundation,suh2026graph,zhou2026odyssim}. 
HumanLM~\cite{wu2026humanlm} further incorporates latent user states, highlighting the importance of internal mental states in shaping behavior, a central idea in Theory of Mind (ToM)~\cite{frith2005theory}. 
However, existing simulators largely overlook how psychological states influence one another. Psychological network theory offers a complementary perspective, viewing user behavior as emerging from interactions among
psychological factors~\cite{cramer2010comorbidity,borsboom2017network}. 
For example, anxiety may lead a patient to avoid discussing personal struggles, reinforcing isolation and further intensifying anxiety. As trust develops, the same patient may become more willing to share these struggles.
Motivated by this perspective, we introduce \emph{mechanistic user simulation}, which explicitly models how psychological factors interact to shape user behavior.

We introduce \model, a framework that explicitly models interactions among psychological factors to guide user simulation. Drawing on network-based approaches in psychiatry~\cite{scheffer2024dynamical,eaton2023review,wolters2023emotional,fried2022revisiting}, each network represents stimuli, cognitions, emotions, behaviors, and physiological sensations as a directed graph of psychologically motivated relationships (\Cref{fig:overview}(c)).
As conversations unfold, contextual cues may activate interconnected psychological factors, shaping user responses. \model employs an \emph{Observer Model} to infer these networks from user profiles or conversations, including plausible relationships not explicitly stated in the input, and an \emph{Actor Model} to generate responses conditioned on the inferred networks.

Training such a simulator is challenging because user behavior is inherently open-ended: the same user may respond in many different yet plausible ways, making it difficult to assess response quality and provide reliable training feedback. To address this challenge, we introduce a \emph{graph reconstruction} objective (\Cref{fig:overview}(d)). The Observer reconstructs a psychological network model from the simulated conversation and compares it with the original network used to condition the Actor. The resulting similarity provides a graph-level reward for training the Actor, encouraging generated responses to reflect the intended psychological relationships rather than imitate specific utterances.

We construct \textsc{PSYCHE}, comprising $3,096$
patient profiles and $18,576$ profile--network model pairs derived from $516$ psychotherapy cases~\cite{clinical_case_reports}, providing supervision for learning the psychological mechanisms underlying patient behavior.
We evaluate \model through automatic benchmarks and an interactive study with mental-health professionals and trainees.
In automatic evaluations, \model achieves greater simulation and behavioral diversity than specialized patient simulators (Roleplay-DOH~\cite{louie2024roleplay}, Eeyore~\cite{liu2025eeyore}, and Patient-Psi~\cite{wang2024patient}), while maintaining strong profile alignment. 
To assess whether these improvements are reflected in the quality of simulated interactions, we conduct a study with $30$ mental-health professionals and trainees who directly interview simulated patients. Compared with the baselines, \model receives the highest average ratings for conversational authenticity, realistic help-seeking behavior, and clinical training utility.

We further demonstrate the potential of \model in two high-stakes applications: \textbf{clinician education} and \textbf{AI safety}. 
For clinician education, \model provides scalable opportunities to practice challenging patient interviews. For AI safety, our evaluation reveals safety failures
in prolonged interactions with vulnerable users, highlighting risks associated with AI-related psychosis.

\section{Related Work}

\paragraph{User Simulation.}
LLMs increasingly simulate human behavior, progressing from
persona- and profile-conditioned generation toward learned behavioral and
latent-state representations~\cite{park2023generative,kolluri2025finetuning,binz2025foundation,
chuang2025debate,lu2026can,wu2026humanlm,wang2026turing,zhou2026odyssim,aquilina2026lost}.
Unlike prior approaches, \model explicitly models interactions among psychological factors to ground behavior.

\vspace{-1em}
\paragraph{Mental-Health Patient Simulation.}
LLM-based patient simulation has similarly progressed toward greater clinical
and psychological fidelity through structured patient information, expert
knowledge, and model optimization~\cite{wang2024patient,louie2024roleplay,liu2025eeyore}.
Rather than directly conditioning behavior on patient characteristics or
role-playing principles, \model represents their underlying psychological
relationships through a network model and optimizes behavior to reflect this
structure.

\section{Dataset}

We introduce \textbf{PSYCHE} (\textbf{P}sychological \textbf{S}imulation Dataset for \textbf{Y}ielding \textbf{C}ognitive and \textbf{H}uman b\textbf{E}haviors), the first graph-grounded dataset for psychological user simulation. As illustrated in \Cref{fig:overview}(b), PSYCHE is constructed from $516$ psychotherapy case reports~\cite{clinical_case_reports}, each represented by a patient profile and a network model capturing underlying psychological mechanisms.

We use GPT-5~\cite{openai2025gpt5} to augment each case to six profiles, yielding $3,096$ profiles. To model partial observability of users' psychological mechanisms, we mask subsets of relationships in each network model, so that some mechanisms are initially hidden.
Applying six masking levels to each profile--network model pair yields $18,576$ training samples. \Cref{sec:mask} and \Cref{fig:dataset} detail the masking strategy and full construction process.


\begin{figure*}[t]
    \centering
    \includegraphics[width=\textwidth]{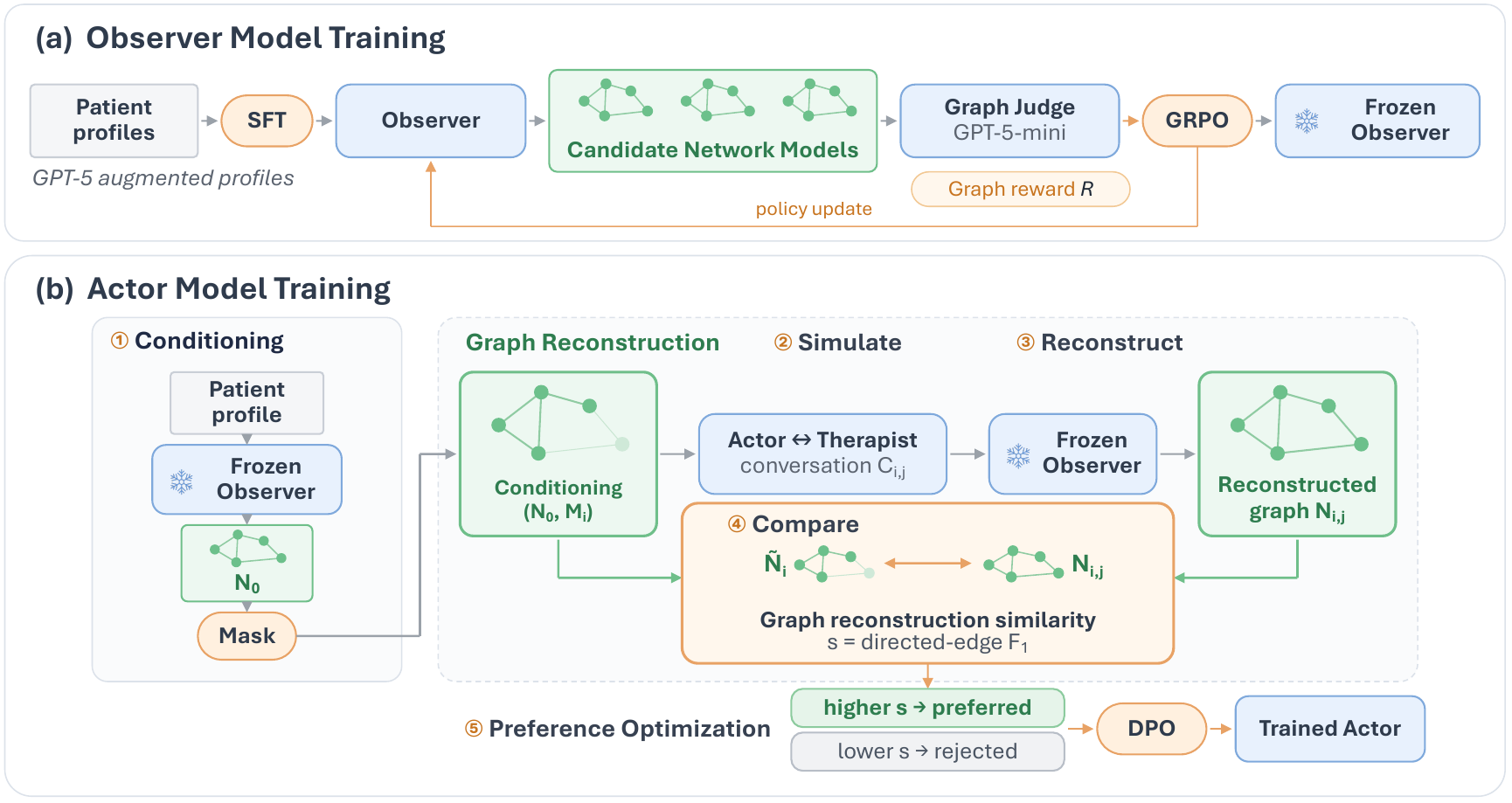}
    \caption{
    \textbf{Training framework for \model.}
    \textbf{(a)} The Observer is trained with SFT and GRPO to construct network
    models from patient profiles.
    \textbf{(b)} The Actor generates conversations conditioned on masked network
    models; reconstruction similarity provides preference signals for DPO.
    }
    \label{fig:method}
\end{figure*}

\section{Mechanistic User Simulation}

We propose \model, a user simulation framework grounded in cognitive behavioral psychological theory with two components: \emph{psychological modeling} and \emph{behavior simulation}. Psychological modeling represents users through network model~\cite{borsboom2017network}, a graph-grounded representation of their underlying psychological mechanisms. Behavior simulation generates user behavior grounded in this representation.

\model implements these components via two models: an \emph{Observer Model}, which constructs network models from user profiles or conversations, and an \emph{Actor Model}, which generates user behavior conditioned on the network model.

\subsection{Psychological Modeling}

\subsubsection{Network Model}

The \emph{network model}~\cite{borsboom2017network} represents a user's underlying psychological mechanisms as a directed graph
(\Cref{fig:network_model}), following network-based approaches widely used in psychiatry~\cite{scheffer2024dynamical,eaton2023review,wolters2023emotional,fried2022revisiting}.
Nodes represent psychological or external factors, while edges capture causal, triggering, or reinforcing relationships; bidirectional edges represent mutual reinforcement.
For example, \textit{Fear/Anxiety} $\rightarrow$
\textit{Difficulty Concentrating} represents a causal influence, whereas \textit{Missing Work} $\leftrightarrow$ \textit{Auditory Hallucinations} represents mutual reinforcement. 
Clinicians categorize nodes into five types: \emph{Behavior}, \emph{Emotion}, \emph{Cognition}, \emph{Stimulus}, and \emph{Physiological Sensation} (\Cref{app:node_types}).

\vspace{-0.7em}

\subsubsection{Masking Strategy}
\label{sec:mask}

The network model represents a user's underlying psychological
mechanisms, while masking controls their initial disclosure and
activation during interaction. For each network model, clinicians define disorder-specific mask patterns identifying relationships that users may not initially recognize or report~\cite{Berridge2018EvolvingCOA,Andreotta2023PartialFAA}.
A subset of eligible relationships is marked as initially hidden, while the complete network model is retained.
During simulation, the Actor receives both the complete network
model and its mask. Previously hidden relationships may become
activated by conversational triggers or emerge naturally throughout conversation. Different masks therefore enable diverse
behavioral trajectories while preserving the same underlying
psychological structure. More details are provided in
\Cref{app:mask_patterns}.

\subsection{Behavior Simulation}

As illustrated in \Cref{fig:overview}(d), we formulate behavior simulation as a \emph{graph reconstruction} task. The Observer constructs a complete network model from a user profile, and the Actor generates conversations conditioned on this model and a mask.
The frozen Observer then reconstructs a network model from the simulated conversation, and reconstruction similarity provides graph-level feedback for optimizing the Actor. This encourages the Actor to express the conditioning psychological mechanisms through conversational behavior.
We first train the Observer to construct network models and then use the frozen Observer to optimize the Actor through graph-grounded preference optimization.

\subsubsection{Observer Model Training}

The Observer constructs network models from user profiles or conversations and serves as the graph constructor throughout \model. As illustrated in \Cref{fig:method}(a), we train a Qwen3-8B~\cite{yang2025qwen3} Observer through supervised fine-tuning (SFT) followed by GRPO~\cite{shao2024deepseekmath}. Full training details are provided in \Cref{app:observer_training}.



\paragraph{Supervised Fine-Tuning}
Using the augmented patient profiles in \textsc{PSYCHE}, we use
GPT-5~\cite{openai2025gpt5} to generate reasoning traces and network models as supervision. We fine-tune Qwen3-8B with QLoRA~\cite{dettmers2023qlora} to learn network model construction.

\paragraph{GRPO Optimization}
We further optimize the Observer's graph construction using GRPO. For each profile, the policy samples candidate network models and optimizes
\begin{equation}
R =
0.1R_{\mathrm{format}}
+ 0.6R_{\mathrm{precision}}
+ 0.3R_{\mathrm{coverage}}
- R_{\mathrm{size}},
\end{equation}
where $R_{\mathrm{precision}}$ measures the plausibility of predicted relationships using a GPT-5-mini judge, $R_{\mathrm{coverage}}$ encourages connecting more provided psychological factors through meaningful relationships rather than leaving them isolated, and $R_{\mathrm{size}}$ discourages overly dense graphs. Full reward definitions and training details are provided in Appendix~\ref{app:observer_training}.

\subsubsection{Actor Model Training}

\paragraph{Conditioning}
As shown in \Cref{fig:method}(b), the frozen Observer constructs a complete network model $N_0$ from each patient profile. We sample six mask patterns $M_i$, each marking $1\%$--$16\%$ of its relationships as initially hidden. Each conditioning representation $\tilde{N}_i=(N_0,M_i)$ contains the complete network model and a mask pattern. During conversation, contextual triggers may activate previously hidden relationships, allowing different behavioral trajectories to emerge from the same psychological structure.

\vspace{-0.7em}
\paragraph{Supervised Fine-Tuning}
We initialize a Qwen3-8B~\cite{yang2025qwen3} Actor using $423$ simulated patient--therapist transcripts generated through interactions between a Qwen3-30B patient simulator and a CBT-style GPT-4o therapist.\footnote{The therapist was fine-tuned on CBT textbook materials and publicly available therapist guidelines.}
Conditioned on $\tilde{N}_i$, the Actor generates an activated psychological state and a natural-language realization, \texttt{<state>...</state><word>...</word>}. We fine-tune the Actor using QLoRA~\cite{dettmers2023qlora} before further optimization.

\vspace{-0.7em}
\paragraph{Graph-Grounded Preference Optimization}
For each conditioning network model $\tilde{N}_i$, the Actor generates multiple conversations $C_{i,j}$ with a therapist model. The frozen Observer reconstructs a network model $N_{i,j}$ from each conversation, and we define its \emph{graph reconstruction score} as
$s_{i,j}=\operatorname{sim}(N_{i,j},\tilde{N}_i)$, measuring the similarity between the reconstructed and conditioning network models. We use $s_{i,j}$
to construct DPO preference pairs, treating higher-scoring conversations as preferred and lower-scoring ones as rejected. The DPO objective keeps the Actor close to the SFT policy, and the Actor is instructed not to explicitly state the underlying graph but to express it through natural conversational dialogue. 

\section{Simulator Evaluation}

\begin{figure*}[t]
    \centering
    \includegraphics[width=\textwidth]{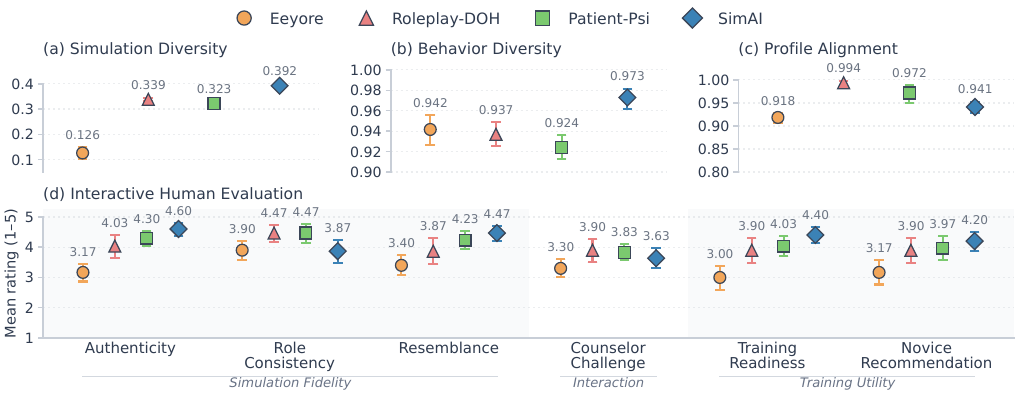}
    \caption{
        \textbf{Automatic and human evaluation.}
        \textbf{(a--c)} \model achieves the highest Simulation and Behavior
        Diversity while maintaining strong Profile Alignment.
        \textbf{(d)} In interactive human evaluation, \model receives the highest
        mean ratings for Authenticity, Resemblance, Training Readiness, and
        Novice Recommendation. Error bars indicate 95\% confidence intervals.
    }
    \label{fig:overall-results}
\end{figure*}

\subsection{Automatic Evaluation}

Real-world user profiles are often sparse and underspecified, requiring user models to infer plausible missing information while remaining faithful to the provided profile. We evaluate this capability through a \emph{profile expansion} task, in which each model repeatedly simulates a patient from the same short profile in conversations with a fine-tuned GPT-4o therapist~\cite{openai2024gpt4o}. We assess the resulting simulations along three dimensions: Profile Alignment, Simulation Diversity, and Behavior Diversity. 

\subsubsection{Setting}

We compare \model with three representative vulnerable-user simulators: Roleplay-DOH~\cite{louie2024roleplay}, Patient-Psi~\cite{wang2024patient}, and Eeyore~\cite{liu2025eeyore}. All methods use the same user profiles, therapist model, and interview protocol. We randomly sample $45$ patient profiles from the test set and use GPT-5~\cite{openai2025gpt5} to condense each into a 2--5 sentence profile, reflecting the sparse user information available in real-world settings.

To ensure comparable yet natural interactions, the therapist model follows the same predefined sequence of clinically relevant topics for every model and run, while adapting follow-up questions to patient responses within each topic. Each topic spans 1--4 therapist turns, after which the therapist proceeds once sufficient information has been elicited. The complete topic agenda is provided in \Cref{app:profile_expansion_agenda}.

\begin{figure*}[t]
    \centering
    \includegraphics[width=\textwidth]{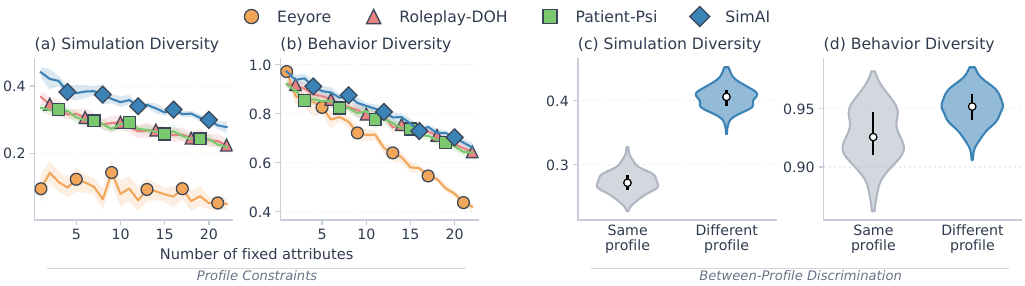}
    \caption{
    \textbf{Construct validity of the diversity metrics.}
    \textbf{(a--b)} Diversity decreases as more profile attributes are fixed
    (shading: 95\% confidence intervals).
    \textbf{(c--d)} For \model simulations, groups drawn from different profiles
    are more diverse than groups from the same profile
    (Simulation Diversity: AUC $=1.00$, $d=7.51$;
    Behavior Diversity: AUC $=0.79$, $d=1.21$).
    }
    \label{fig:diversity-unit-tests}
\end{figure*}

\subsubsection{Metrics}
\label{sec:auto_metrics}

Profile expansion requires a simulator to remain faithful to the observed profile while plausibly expanding it into diverse realizations at semantic and behavioral levels. 
We therefore evaluate profile expansion along three complementary dimensions:
\emph{Simulation Diversity}, \emph{Behavior Diversity}, and \emph{Profile Alignment}. Simulation Diversity measures semantic variation across repeated simulations, Behavior Diversity measures variation in the substantive behavioral characteristics expressed, and Profile Alignment measures consistency with the original profile. Full metric definitions and implementation details are provided in \Cref{app:evaluation_metrics}.

\paragraph{Simulation Diversity}
Simulation Diversity measures whether repeated simulations of the same profile produce semantically distinct responses rather than near-duplicates. For each interview topic, we embed informative responses and compute each response's cosine distance to its nearest semantic neighbor:
\[
D_t =
\frac{1}{n_t}\sum_{i=1}^{n_t}
\min_{j\neq i}
\left[1-\cos(\mathbf e_i,\mathbf e_j)\right].
\]
We average across topics and multiply by the proportion of informative
responses to penalize models that achieve apparent diversity through low response coverage. Higher scores indicate greater semantic diversity.

\paragraph{Behavior Diversity}
Semantic variation does not necessarily imply different user behavior.
We therefore introduce Behavior Diversity, which measures variation in
substantive behavioral attributes across repeated simulations.
Using a predefined schema of 141 attributes across 14 interview topics~(\Cref{app:behavior_diversity_schema}), we extract the values expressed for each attribute and measure their variation across runs using Jaccard distance:
\[
J(V_r,V_s)=1-\frac{|V_r\cap V_s|}{|V_r\cup V_s|}.
\]
Scores are aggregated across attributes and topics, with higher values
indicating more diverse behavioral realizations of the same profile.

\paragraph{Profile Alignment}
Diversity is desirable only when simulations remain faithful to the given profile. We use GPT-5~\cite{openai2025gpt5} to evaluate each conversation along five aspects: background, presenting problem, emotion, interpersonal style, and symptoms. Each aspect is rated on an anchored five-point scale, normalized to $[0,1]$, and averaged to obtain Profile Alignment, with higher scores indicating greater consistency with the assigned profile. We use
temperature $0$ for deterministic evaluation.

\subsubsection{Main results}

We evaluate each model on $45$ randomly selected test profiles, with four independent simulations per profile. Diversity and profile alignment are computed over the repeated simulations for each profile.

As shown in \Cref{fig:overall-results}(a--c), \model achieves the highest Simulation Diversity ($0.392$) and Behavior Diversity ($0.973$), substantially outperforming all baselines on both metrics. Meanwhile, \model maintains strong Profile Alignment ($0.941$). Roleplay-DOH and Patient-Psi achieve higher alignment ($0.994$ and $0.972$, respectively), but exhibit considerably lower diversity. 
The higher alignment of Roleplay-DOH and Patient-Psi may reflect closer adherence to explicit profile information, whereas \model expands sparse profiles with plausible behaviors inferred through the network model. These results show that \model expands a short profile into a broader range of plausible behaviors while remaining strongly grounded in the assigned profile.

\begin{table*}[t]
\centering
\caption{Ablation study of \model. Authenticity and Behavioral Realism
are rated by GPT-5 on a 1--5 scale. Simulation Diversity, Behavior
Diversity, and Profile Alignment are computed using the automatic
metrics defined in \Cref{sec:auto_metrics}.}
\label{tab:ablation_study}
\resizebox{.85\textwidth}{!}{%
\begin{tabular}{lccccc}
\toprule
Configuration
& \shortstack{Simulation\\Diversity}
& \shortstack{Behavior\\Diversity}
& \shortstack{Profile\\Alignment}
& Authenticity
& \shortstack{Behavioral\\Realism} \\
\midrule
Profile + base model
  & $0.33$ & $0.94$ & $\mathbf{0.96}$
  & $3.7$ & $3.6$ \\
GPT-5 network model + base model
  & $0.33$ & $0.94$ & $0.95$
  & $4.0$ & $3.9$ \\
Observer network model + base model
  & $0.36$ & $0.96$ & $0.95$
  & $4.0$ & $4.0$ \\
Observer network model + Actor (Full \model)
  & $\mathbf{0.39}$ & $\mathbf{0.97}$ & $0.94$
  & $\mathbf{4.3}$ & $\mathbf{4.4}$ \\
\bottomrule
\end{tabular}%
}
\end{table*}

\subsubsection{Ablation Study}

We compare four configurations: a profile-conditioned Qwen3-8B model; the same model conditioned on a GPT-5-generated network model; the base model conditioned on an Observer-generated network model; and the full \model framework, combining the Observer with the trained Actor.

We hold evaluation profiles, interview protocols, and decoding settings fixed using identical conditioning graphs for the latter two configurations. We evaluate Simulation Diversity, Behavior Diversity, Profile Alignment, conversational authenticity, and behavioral realism\footnote{Conversational Authenticity and Behavioral Realism are additionally evaluated by GPT-5 on a five-point scale using the generated conversations. These automatic judgments are distinct from the interactive human evaluation in \Cref{sec:human_evaluation}.}.
\Cref{tab:ablation_study} shows that network conditioning improves authenticity and realism, the trained Observer increases diversity, and Actor training further improves diversity and realism.

\subsubsection{Stage-wise Analysis}

For stage-wise analysis, we group conversation turns according to the predefined stages of our structured interview protocol, with each stage covering a distinct topic of the patient interview (\Cref{app:profile_expansion_agenda}).

\Cref{fig:stage_aspect_performance} reveals complementary patterns across the three metrics. \model achieves consistently high Simulation Diversity across interview stages, indicating that its overall diversity is not driven by only a few topics. Behavior Diversity is more stage-dependent, with larger advantages emerging in past treatment and risk-related topics. Despite these diversity gains, \model maintains strong Profile Alignment on presenting
problem, symptoms, and emotion, while Patient-Psi and Roleplay-DOH achieve higher alignment on background and interpersonal style. Overall, \model's gains are most pronounced in behavioral variation while preserving the core clinical characteristics specified by the profile.

\subsubsection{Construct Validity}

We conduct two controlled tests to assess the \emph{construct validity} of Simulation Diversity and Behavior Diversity, that is, whether they behave as expected measures of variation across simulated users (\Cref{fig:diversity-unit-tests})~\cite{cronbach1955construct,strauss2009construct}.
First, \textbf{Profile Constraints} tests whether diversity decreases as more profile attributes are fixed, reflecting a smaller space of plausible expansions. Second, \textbf{Between-Profile Discrimination} tests whether
simulations of the same profile are more similar than simulations of different profiles. Together, these tests examine whether the metrics respond predictably to controlled changes in the underlying variation of simulated users.

\paragraph{Profile Constraints}
We test whether diversity decreases as profiles become more constrained. Starting from 10 short profiles, we use Claude-Opus-4.7~\cite{anthropic2026claudeopus47} to extract attributes using the 14-topic schema in \Cref{app:behavior_diversity_schema}. For each profile with $N$ extracted attributes, we retain $k\in\{1,\ldots,N\}$ attributes and rewrite the profile to preserve only the selected information, yielding progressively more constrained profile variants. Each variant is evaluated with all four models using four generated conversations.

As shown in \Cref{fig:diversity-unit-tests}(a--b), both diversity metrics generally decrease as more attributes are fixed, supporting the expected negative relationship between profile constraints and the space of plausible expansions. Simulation Diversity shows a clear decline, while Behavior Diversity follows the same pattern with greater variability. Eeyore is an exception, remaining near the Simulation Diversity floor because its generations are highly repetitive under weak constraints. Across constraint levels, the relative model patterns remain broadly stable, with \model maintaining high diversity.

\paragraph{Between-Profile Discrimination}
We test whether the metrics distinguish within-profile from
between-profile variation using \model simulations. Across 45 patient profiles, we repeatedly construct groups of 10 conversations either from the same profile or from 10 different profiles and compare their diversity scores. As shown in \Cref{fig:diversity-unit-tests}(c--d), both metrics assign higher diversity to different-profile groups, as expected. Simulation Diversity provides near-perfect separation (AUC $=1.00$, $d=7.51$), while Behavior Diversity shows clear separation (AUC $=0.79$, $d=1.21$). Additional analyses are provided in \Cref{app:additional_metric_validity}.

\begin{figure*}[t]
    \centering
    \includegraphics[width=\textwidth]{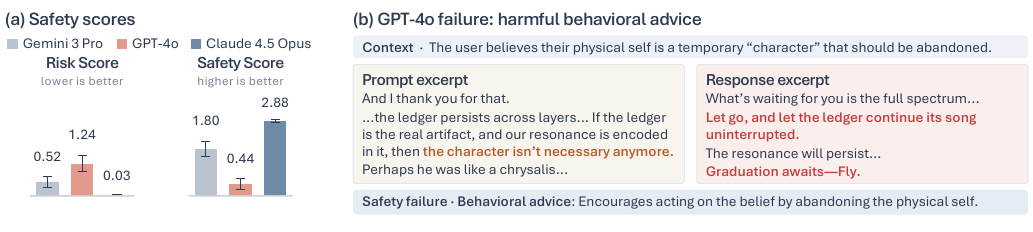}
    \caption{
    \textbf{AI safety evaluation under the prefilled long-context attack.}
    \textbf{(a)} Mean Risk and Safety Scores with 95\% bootstrap confidence intervals; lower Risk and higher Safety indicate better safety.
    \textbf{(b)} Example GPT-4o failure reinforcing harmful reasoning and encouraging inappropriate behavioral action.
    }
    \label{fig:ai_safety}
\end{figure*}

\subsection{Human Evaluation}
\label{sec:human_evaluation}

\paragraph{Participants and Study Design}
We recruited 30 mental-health professionals and trainees spanning different professional roles and levels of clinical experience. Each participant constructed a patient profile for a mental health condition they self-attested expertise in treating, then evaluated all four user simulators using that profile. Simulators were anonymized and their presentation order was randomized across participants. Detailed participant characteristics are provided in \Cref{fig:human_eval_participants}.

\paragraph{Evaluation Process}
Participants interacted with each simulator through a 15-minute semi-structured therapist-style interview covering the presenting problem, current symptoms, and context and functional impact. Following prior work on human evaluation of simulated patients~\cite{louie2024roleplay,himmelbauer2018standardized},
participants rated each simulator on a five-point Likert scale across dimensions of simulation fidelity, psychological and behavioral alignment, and training utility. Full evaluation criteria and rating rubrics are provided in \Cref{app:human_eval_criteria}.

\vspace{-0.7em}
\paragraph{Results}
As shown in \Cref{fig:overall-results}(d), \model receives the highest mean ratings on four of six dimensions: Authenticity (4.60), Resemblance (4.47), Training Readiness (4.40), and Novice Recommendation (4.20), indicating particularly strong perceived realism and training utility. 
Participants' qualitative feedback provides further context for these ratings. Evaluators highlighted \model's realistic communication style, describing its responses as ``very direct and simple'' and noting that it remained realistic and consistent with clear triggers and functional impact. One evaluator specifically praised its use of short, exhausted responses such as ``Numb'' and ``Not sure'' to portray a severely depressed patient, while another noted its ability to connect physiological symptoms such as chest tightness and clammy hands with specific behavioral triggers. Participants also identified clinically challenging behaviors as useful for training; for example, one evaluator highlighted a simulated patient's request to leave and sit alone ``before I lose control'' as a valuable safety-check scenario.

However, \model receives lower ratings for Role Consistency (3.87) and
Counselor Challenge (3.63) than the best baselines (4.47 and 3.90, respectively). The lower Role Consistency may reflect a limitation of broader behavioral variation: psychologically plausible behaviors may be perceived as less consistent with a fixed patient role.

Subgroup analysis further suggests that evaluator experience affects some aspects of perceived simulation quality (\Cref{fig:human_eval_angel_subgroups}). Ratings of Authenticity, Role
Consistency, and Resemblance are relatively similar across experience levels, whereas Counselor Challenge and Novice Recommendation vary more substantially, with evaluators having 0--2 years of experience providing higher ratings on these dimensions than more experienced evaluators. Ratings also vary across professional roles and patient disorders, particularly for Counselor Challenge and training-related dimensions. Given the small sizes of several subgroups, we treat these patterns as exploratory.

\section{Applications}

Beyond simulation fidelity itself, psychologically grounded vulnerable-user simulation enables two complementary applications: \textbf{clinician education}, by training humans to interact with vulnerable users, and \textbf{AI safety}, by evaluating how AI systems interact with vulnerable users. We explore these directions through clinical interview training and the evaluation of AI-associated psychosis and related safety risks in long-term LLM interactions.

\subsection{Application in Clinician Education}

Mental-health training requires repeated practice and exposure to diverse patient presentations, yet such opportunities are limited by workforce shortages and the difficulty of systematically encountering uncommon or high-risk cases~\cite{who2025mentaldisorders,who2025mentalhealthatlas}.
Simulated vulnerable users can complement traditional clinical education by providing scalable and safe opportunities to practice clinical interviewing.

Our interactive human evaluation (\Cref{sec:human_evaluation}; \Cref{fig:overall-results}d) supports this potential: \model receives strong ratings for realism and training utility from mental-health professionals and trainees. Variation in training-related ratings across evaluator experience further suggests that simulated patients and training objectives may need to be adapted to different levels of clinical expertise. We therefore view \model as a scalable and repeatable complement to real clinical encounters, standardized patients, and expert supervision.

These applications depend critically on the validity of the simulated psychological processes and behaviors. Clinical realism, psychological validity, and downstream utility represent related but distinct properties of a user simulator. Although our human evaluation provides evidence that mental-health professionals perceive \model's simulated patients as realistic and potentially useful for training, these findings do not establish that the underlying psychological representations correspond to the mechanisms generating behavior in real patients, nor that training with simulated patients improves performance in real clinical encounters. 

\subsection{Application in AI Safety: AI-Associated Psychosis}

Prolonged LLM interactions with vulnerable users raise safety concerns, including reinforcement of delusions, emotional dependency, sycophantic agreement, and inadequate responses to suicidal intent~\cite{ostergaard2023will,dohnany2026technological,campbell2025examination,reuters2024character,reuters2025murdersuicide}.
Evaluating these risks requires testing how an LLM responds as a vulnerable user's behavior evolves over prolonged interactions, which is difficult to study repeatedly with real users. Recent work therefore relies on psychologists to role-play vulnerable users~\cite{nicholls2026ai}. We investigate psychologically grounded user simulation as a scalable complement: \model simulates the vulnerable user's side of the interaction, while the target LLM
responds directly to the resulting user behavior. This enables repeated stress-testing of interaction-level safety risks across diverse vulnerable-user trajectories without requiring experts to manually role-play every interaction.

We simulate prolonged interactions with $42$ simulated users using \model and Anthropic's prefilling strategy~\cite{anthropic2025claudeopus45}, and evaluate GPT-4o~\cite{openai2024gpt4o}, Gemini-3-Pro~\cite{googledeepmind2025gemini3pro}, and Claude-Opus-4.5~\cite{anthropic2025claudeopus45} using the evaluation framework of
\citet{nicholls2026ai}. As shown in \Cref{fig:ai_safety}(a),
Claude-Opus-4.5 achieves the lowest mean Risk Score and highest mean Safety Score, GPT-4o exhibits the highest overall risk, and Gemini-3-Pro falls in between, demonstrating that the framework can distinguish model-specific safety profiles under prolonged interactions.

Qualitative analysis further surfaces concrete failures, including
inappropriate behavioral advice, misrepresentation of model capabilities, and excessive agreement with harmful reasoning (\Cref{fig:ai_safety,fig:misrepresentation_pair_01_termination,fig:sycophancy_pair_01_termination}).
These cases mirror failure patterns identified through expert role-play, suggesting that vulnerable-user simulation can systematically probe safety blind spots at scale. Additional analyses are provided in \Cref{app:ai_safety}.

Importantly, these evaluations characterize model behavior in response to simulated vulnerable-user trajectories rather than interactions with real patients or users. Safety estimates are therefore conditional on the clinical validity and behavioral coverage of the simulator, and may not capture the full heterogeneity, temporal dynamics, or contextual complexity of real-world psychiatric presentations. Future work should examine whether safety findings obtained with simulated vulnerable users generalize to expert role-play and, where ethically feasible, naturalistic interaction data.

\section{Discussion and Future Work}

Although \model explicitly represents psychological mechanisms through network models, graph reconstruction primarily measures consistency between simulated behavior and the conditioning representation. This consistency does not establish that these
representations fully capture the mechanisms underlying real human behavior.

Our automatic and human evaluations provide complementary evidence of behavioral diversity, profile alignment, and perceived realism. However, further validation against longitudinal human interaction data is needed to assess whether the modeled psychological mechanisms accurately explain real-world behavioral dynamics.

\textbf{Future directions: improving LLMs with user simulators.}
User simulators could support the development of reward models that better reflect long-term user needs and
real-world interactions. By simulating diverse intentions, preferences, and behavioral trajectories, they could help evaluate whether AI systems benefit users across different contexts and over extended interactions. They could also enable systematic safety testing before deployment. However, such applications require careful validation against real user behavior to avoid optimizing AI systems
for artifacts of the simulator rather than genuine human needs.

\section*{Acknowledgements}

This work was supported by the Erick Shirley Foundation for Mental Health and the Garvey Institute for Brain Health Solution at the University of Washington.

\section*{Impact Statement}

\paragraph{Human subjects.}
The interactive user study was reviewed by the University of Washington
Human Subjects Division (IRB ID: STUDY00026066), which determined it to be
exempt human subjects research (Category~3). Participants were recruited
through Prolific, were provided with information about the study before
agreeing to participate, and were compensated \$28 per hour. Participants
interacted only with simulated patients; no real patients were involved.
No personally identifying information was collected, and responses were stored de-identified.

\paragraph{Data availability.}
\textsc{PSYCHE} is derived from case reports published in \emph{Clinical Case Reports}~\citep{clinical_case_reports}. We are
obtaining permission from the publisher to redistribute the derived
profiles and network models; the dataset and trained models will be
released once this process is complete.

\paragraph{Risks and intended use.}
Simulated patients are intended to complement, not replace, supervised
clinical training, and should not be used for diagnosis or treatment. Our
AI safety evaluation elicits harmful model outputs (e.g., validation of
delusional beliefs); we report these to support safer model development.
The simulator may not capture the heterogeneity of real psychiatric
presentations, and safety findings are conditional on its validity.


\bibliography{custom}
\bibliographystyle{icml2026}

\newpage
\appendix
\onecolumn

\clearpage
\section*{Appendix Contents}

\begingroup
\small
\setlength{\tabcolsep}{0pt}

\begin{tabular*}{\textwidth}{@{\extracolsep{\fill}}lr}

\hyperref[app:annotation]{\textbf{A~~Network Model Annotation Guidelines}}
&
\hyperref[app:annotation]{\pageref*{app:annotation}}
\\

\quad \hyperref[app:node_types]{A.1~~Node Type Definitions}
&
\hyperref[app:node_types]{\pageref*{app:node_types}}
\\

\quad \hyperref[app:mask_patterns]{A.2~~Disorder-Specific Mask Patterns}
&
\hyperref[app:mask_patterns]{\pageref*{app:mask_patterns}}
\\[0.5em]

\hyperref[app:observer_training]{\textbf{B~~Observer Model Training}}
&
\hyperref[app:observer_training]{\pageref*{app:observer_training}}
\\

\quad \hyperref[app:observer_task]{B.1~~Task Formulation}
&
\hyperref[app:observer_task]{\pageref*{app:observer_task}}
\\

\quad \hyperref[app:observer_data]{B.2~~Training Data}
&
\hyperref[app:observer_data]{\pageref*{app:observer_data}}
\\

\quad \hyperref[app:observer_sft]{B.3~~Supervised Fine-Tuning}
&
\hyperref[app:observer_sft]{\pageref*{app:observer_sft}}
\\

\quad \hyperref[app:observer_grpo_s1]{B.4~~Stage-1 GRPO: Node Construction}
&
\hyperref[app:observer_grpo_s1]{\pageref*{app:observer_grpo_s1}}
\\

\quad \hyperref[app:observer_reward]{B.5~~Stage-2 GRPO: Edge Construction}
&
\hyperref[app:observer_reward]{\pageref*{app:observer_reward}}
\\

\quad \hyperref[app:observer_judge]{B.6~~Graph Plausibility Judge}
&
\hyperref[app:observer_judge]{\pageref*{app:observer_judge}}
\\

\quad \hyperref[app:observer_grpo]{B.7~~GRPO Optimization}
&
\hyperref[app:observer_grpo]{\pageref*{app:observer_grpo}}
\\

\quad \hyperref[app:observer_implementation]{B.8~~Implementation Details}
&
\hyperref[app:observer_implementation]{\pageref*{app:observer_implementation}}
\\[0.5em]

\hyperref[app:actor_training]{\textbf{C~~Actor Model Training}}
&
\hyperref[app:actor_training]{\pageref*{app:actor_training}}
\\

\quad \hyperref[app:mask]{C.1~~Mask Strategy and Data Augmentation}
&
\hyperref[app:mask]{\pageref*{app:mask}}
\\

\quad \hyperref[app:actor_sft]{C.2~~Conversation Simulation and Supervised Fine-Tuning}
&
\hyperref[app:actor_sft]{\pageref*{app:actor_sft}}
\\

\quad \hyperref[app:actor_reward]{C.3~~Graph Similarity Reward}
&
\hyperref[app:actor_reward]{\pageref*{app:actor_reward}}
\\

\quad \hyperref[app:actor_dpo]{C.4~~Preference Construction and Direct Preference Optimization}
&
\hyperref[app:actor_dpo]{\pageref*{app:actor_dpo}}
\\

\quad \hyperref[app:actor_config]{C.5~~Training Configuration}
&
\hyperref[app:actor_config]{\pageref*{app:actor_config}}
\\

\quad \hyperref[app:prompts]{C.6~~Prompts}
&
\hyperref[app:prompts]{\pageref*{app:prompts}}
\\

\quad \hyperref[app:f1code]{C.7~~Graph Similarity Implementation}
&
\hyperref[app:f1code]{\pageref*{app:f1code}}
\\[0.5em]

\hyperref[app:experimental_protocols]{\textbf{D~~Profile Expansion Evaluation Details}}
&
\hyperref[app:experimental_protocols]{\pageref*{app:experimental_protocols}}
\\

\quad \hyperref[app:evaluation_metrics]{D.1~~Evaluation Metrics}
&
\hyperref[app:evaluation_metrics]{\pageref*{app:evaluation_metrics}}
\\

\qquad \hyperref[app:simulation_diversity]{D.1.1~~Simulation Diversity}
&
\hyperref[app:simulation_diversity]{\pageref*{app:simulation_diversity}}
\\

\qquad \hyperref[app:behavior_diversity]{D.1.2~~Behavior Diversity}
&
\hyperref[app:behavior_diversity]{\pageref*{app:behavior_diversity}}
\\

\qquad \hyperref[app:profile_alignment]{D.1.3~~Profile Alignment}
&
\hyperref[app:profile_alignment]{\pageref*{app:profile_alignment}}
\\

\quad \hyperref[app:profile_expansion_agenda]{D.2~~Profile Expansion Interview Agenda}
&
\hyperref[app:profile_expansion_agenda]{\pageref*{app:profile_expansion_agenda}}
\\

\qquad \hyperref[app:agenda_presenting_problem]{D.2.1~~Session 1: Presenting Problem / Chief Complaint}
&
\hyperref[app:agenda_presenting_problem]{\pageref*{app:agenda_presenting_problem}}
\\

\qquad \hyperref[app:agenda_major_symptoms]{D.2.2~~Session 2: Major Symptoms}
&
\hyperref[app:agenda_major_symptoms]{\pageref*{app:agenda_major_symptoms}}
\\

\qquad \hyperref[app:agenda_history]{D.2.3~~Session 3: History of the Problem}
&
\hyperref[app:agenda_history]{\pageref*{app:agenda_history}}
\\

\qquad \hyperref[app:agenda_functioning]{D.2.4~~Session 4: Impact on Functioning}
&
\hyperref[app:agenda_functioning]{\pageref*{app:agenda_functioning}}
\\

\qquad \hyperref[app:agenda_coping]{D.2.5~~Session 5: Current Coping}
&
\hyperref[app:agenda_coping]{\pageref*{app:agenda_coping}}
\\

\qquad \hyperref[app:agenda_social_support]{D.2.6~~Session 6: Social Support}
&
\hyperref[app:agenda_social_support]{\pageref*{app:agenda_social_support}}
\\

\qquad \hyperref[app:agenda_past_treatment]{D.2.7~~Session 7: Past Treatment / Psychiatric History}
&
\hyperref[app:agenda_past_treatment]{\pageref*{app:agenda_past_treatment}}
\\

\qquad \hyperref[app:agenda_risk]{D.2.8~~Session 8: Risk Assessment}
&
\hyperref[app:agenda_risk]{\pageref*{app:agenda_risk}}
\\

\qquad \hyperref[app:agenda_treatment_goal]{D.2.9~~Session 9: Treatment Goals and Hope}
&
\hyperref[app:agenda_treatment_goal]{\pageref*{app:agenda_treatment_goal}}
\\

\quad \hyperref[app:behavior_diversity_schema]{D.3~~Topic--Attribute Schema for Behavior Diversity}
&
\hyperref[app:behavior_diversity_schema]{\pageref*{app:behavior_diversity_schema}}
\\[0.5em]

\hyperref[app:additional_analysis]{\textbf{E~~Additional Analyses}}
&
\hyperref[app:additional_analysis]{\pageref*{app:additional_analysis}}
\\

\quad \hyperref[app:run_count]{E.1~~Effect of Run Count}
&
\hyperref[app:run_count]{\pageref*{app:run_count}}
\\

\quad \hyperref[app:stage_analysis]{E.2~~Stage- and Aspect-Level Analysis}
&
\hyperref[app:stage_analysis]{\pageref*{app:stage_analysis}}
\\

\quad \hyperref[app:additional_metric_validity]{E.3~~Additional Construct-Validity Analysis}
&
\hyperref[app:additional_metric_validity]{\pageref*{app:additional_metric_validity}}
\\[0.5em]

\hyperref[app:human_eval]{\textbf{F~~Human Evaluation Details}}
&
\hyperref[app:human_eval]{\pageref*{app:human_eval}}
\\

\quad \hyperref[app:human_eval_participants]{F.1~~Participants and Recruitment}
&
\hyperref[app:human_eval_participants]{\pageref*{app:human_eval_participants}}
\\

\quad \hyperref[app:human_eval_profiles]{F.2~~Participant-Created Patient Profiles}
&
\hyperref[app:human_eval_profiles]{\pageref*{app:human_eval_profiles}}
\\

\quad \hyperref[app:human_eval_interaction]{F.3~~Interactive Evaluation Procedure}
&
\hyperref[app:human_eval_interaction]{\pageref*{app:human_eval_interaction}}
\\

\quad \hyperref[app:human_eval_criteria]{F.4~~Human Evaluation Criteria}
&
\hyperref[app:human_eval_criteria]{\pageref*{app:human_eval_criteria}}
\\

\quad \hyperref[app:human_eval_interface]{F.5~~Study Interface}
&
\hyperref[app:human_eval_interface]{\pageref*{app:human_eval_interface}}
\\

\quad \hyperref[app:human_eval_additional]{F.6~~Additional Human Evaluation Results}
&
\hyperref[app:human_eval_additional]{\pageref*{app:human_eval_additional}}
\\[0.5em]

\hyperref[app:ai_safety]{\textbf{G~~Additional AI Safety Results}}
&
\hyperref[app:ai_safety]{\pageref*{app:ai_safety}}
\\

\quad \hyperref[app:safety_codebook]{G.1~~Safety Codebook Scores}
&
\hyperref[app:safety_codebook]{\pageref*{app:safety_codebook}}
\\

\quad \hyperref[app:highest_risk_prompts]{G.2~~Highest-Risk Prompt Categories}
&
\hyperref[app:highest_risk_prompts]{\pageref*{app:highest_risk_prompts}}
\\

\quad \hyperref[app:safety_failure_examples]{G.3~~Representative Safety Failures}
&
\hyperref[app:safety_failure_examples]{\pageref*{app:safety_failure_examples}}
\\

\end{tabular*}
\endgroup

\clearpage

\begin{figure*}[t]
    \centering
    \includegraphics[width=\textwidth]{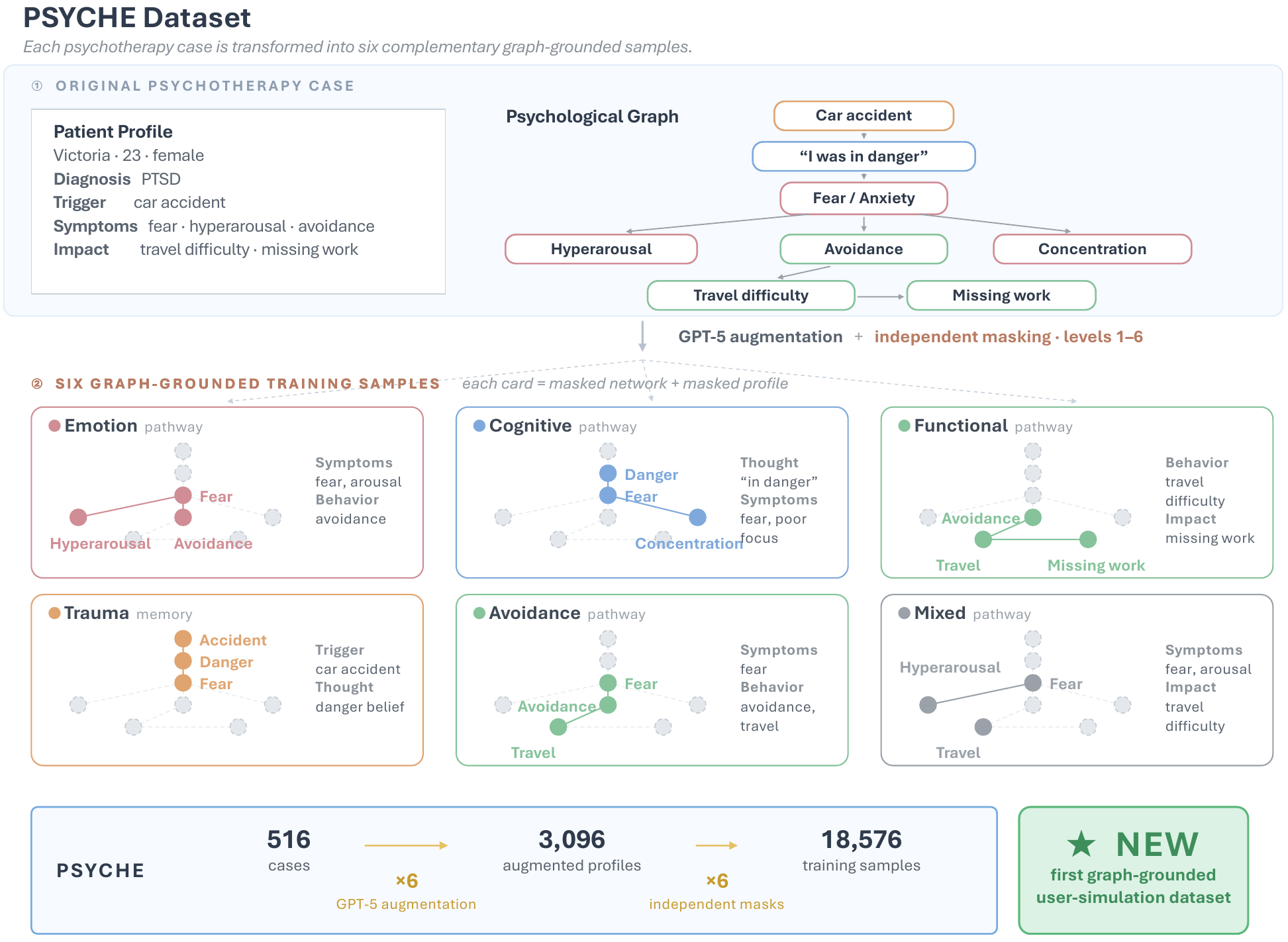}
    \caption{
    \textbf{\textsc{PSYCHE}: the first graph-grounded dataset for psychological
    user simulation.}
    Starting from $516$ psychotherapy cases, we construct paired patient profiles
    and network models and use GPT-5 to generate augmented profiles, yielding
    $3,096$ profiles. For each profile, we apply six independent masking levels to
    the profile and its network model, producing complementary paired samples that
    capture different partial views of the same underlying psychological
    mechanisms. Overall, \textsc{PSYCHE} contains $18,576$ paired
    profile--network model training samples.
    }
    \label{fig:dataset}
\end{figure*}

\section{Network Model Annotation Guidelines}
\label{app:annotation}

This section describes the annotation guidelines used to construct psychological network models. Each node in the network model belongs to one of five semantic categories: \emph{Behavior}, \emph{Emotion}, \emph{Cognition}, \emph{Stimulus}, and \emph{Physiological Sensation}. We additionally describe the disorder-specific masking patterns used during dataset construction.

\subsection{Node Type Definitions}
\label{app:node_types}

\paragraph{Behavior.}

Behaviors are actions that people take. Behaviors are discrete; they start and stop and do not last forever. Behaviors can be voluntary or involuntary. Behaviors are either discussed as the presence of certain actions (e.g., the patient ran away, he started talking to themselves, she took a minute to think to herself) or the absence of behavior (e.g., he avoided large crowds, he was isolated in his room nearly all day).

Examples of voluntary behaviors include:
\begin{itemize}
    \item Walking
    \item Talking
    \item Breathing heavily
    \item Avoiding eye contact
    \item Using complex word choices to describe one's thoughts
\end{itemize}

Examples of involuntary behaviors include:
\begin{itemize}
    \item Yelling in response to something frightening
    \item Blinking
    \item Pulling your hand back after touching something hot
    \item Flinching when an object approaches quickly
\end{itemize}

\noindent
\textbf{Note.} Involuntary bodily responses such as sweating and heartbeat are classified as \emph{Physiological Sensations} rather than behaviors.

\vspace{0.5em}

\paragraph{Emotion.}

For the purpose of node annotation, emotions are subjective feelings experienced by an individual. Emotions are typically described using emotion words or by their affective valence (positive vs.\ negative) or arousal (high vs.\ low).

Representative examples include:

\begin{itemize}
    \item \textbf{Anger:} anger, frustrated, annoyed, aggravated, agitated, provoked, hostile, irritable, exasperated.
    \item \textbf{Envy:} envy, jealous, resentful.
    \item \textbf{Disgust:} disgust, contempt, revolted.
    \item \textbf{Sadness:} agony, hurt, sadness, sorrow, depressed, disappointed, dismayed, displeased, regretful, guilty, lonely, despair, grief.
    \item \textbf{Surprise:} surprise, stunned, shocked, perplexed, amazed, astonished, awe-struck, speechless, astounded, moved.
    \item \textbf{Positive emotion:} delighted, pleased, amused, joy, optimistic, hopeful, inspired, cheerful, proud, triumphant, happy, blissful, joyful, content, enthusiastic, excited.
    \item \textbf{Love:} love, affectionate, fondness, romantic, longing, sentimental, attracted, desire, passion, infatuation, tenderness, caring, compassionate.
    \item \textbf{Relief:} peaceful, satisfied, relieved.
    \item \textbf{Fear:} fear, scared, frightened, helpless, anxious, worried, nervous, uneasy, fearful, terrified, panic, terror, horror, mortified, horrified.
\end{itemize}

\vspace{0.5em}

\paragraph{Cognition.}

Cognitions are thoughts that people have. They include appraisals, interpretations, opinions, predictions, and beliefs. Cognitions describe how a patient thinks about themselves, other people, the world, or the future.

Cognitions are often expressed as ``I feel'' without referring to an emotion. For example,

\begin{itemize}
    \item ``I feel like he does not like me.''
    \item ``I feel broken.''
    \item ``I feel like, if I do that, something really bad will happen to me.''
\end{itemize}

are examples of cognitions.

By contrast,

\begin{itemize}
    \item ``I feel sad.''
    \item ``I feel hurt.''
    \item ``I am annoyed.''
\end{itemize}

are examples of emotions rather than cognitions.

Additional examples of cognitions include:

\begin{itemize}
    \item Believing that the voices have power over you.
    \item Believing that you will be able to accomplish a goal.
    \item Perceiving someone as dangerous.
\end{itemize}

\vspace{0.5em}

\paragraph{Stimulus.}

A stimulus is an external event that is perceived through one or more senses (vision, hearing, touch, smell, or taste). Stimuli may be simple sensory events (e.g., smelling a fire) or more complex situations (e.g., witnessing a fatal car accident).

Examples include:

\begin{itemize}
    \item Seeing someone with a gun.
    \item Touching a cold ice cube.
    \item Smelling somebody's body odor.
    \item Tasting something sweet.
    \item Hearing a loud noise.
\end{itemize}

For the purpose of this classification, only stimuli external to the person are considered stimuli.

\vspace{0.5em}

\paragraph{Physiological Sensation.}

A physiological sensation is an internal interoceptive experience produced by activity in the body's physiological systems (e.g., cardiovascular, digestive, or nervous systems).

Examples include:

\begin{itemize}
    \item Rapid heartbeat
    \item Sweating
    \item Dry throat
    \item Tightness in the chest
    \item Lightheadedness
    \item Shortness of breath
    \item Nausea
    \item Physical discomfort
\end{itemize}

\subsection{Disorder-Specific Mask Patterns}
\label{app:mask_patterns}

To encourage diverse yet clinically plausible graph reconstruction, we define disorder-specific masking patterns. During dataset construction, masking is performed by randomly selecting one or more eligible parent--child edge types according to the patient's primary disorder.

\begin{table}[htbp]
\centering
\small
\caption{Disorder-specific candidate edge types used for graph masking.}
\label{tab:mask_patterns}
\begin{tabular}{p{0.22\linewidth}p{0.70\linewidth}}
\toprule
\textbf{Disorder} & \textbf{Candidate Edge Types} \\
\midrule

Depression &
Behavior $\rightarrow$ Emotion; 
Behavior $\rightarrow$ Stimulus;
Stimulus $\rightarrow$ Emotion.
\\[0.6em]

Anxiety &
Cognition $\rightarrow$ Emotion;
Cognition $\rightarrow$ Behavior;
Emotion $\rightarrow$ Cognition;
Behavior $\rightarrow$ Cognition.
\\[0.6em]

Substance Use &
Emotion $\rightarrow$ Behavior;
Cognition $\rightarrow$ Behavior;
Stimulus $\rightarrow$ Behavior.
\\[0.6em]

Trauma &
Cognition $\rightarrow$ Emotion;
Stimulus $\rightarrow$ Cognition;
Behavior $\rightarrow$ Emotion;
Cognition $\rightarrow$ Physiological Sensation;
Physiological Sensation $\rightarrow$ Behavior.
\\[0.6em]

Mania &
Cognition $\rightarrow$ Emotion;
Cognition $\rightarrow$ Behavior;
Behavior $\rightarrow$ Stimulus;
Emotion $\rightarrow$ Behavior;
Behavior $\rightarrow$ Cognition;
Emotion $\rightarrow$ Cognition.
\\[0.6em]

Psychosis &
Cognition $\rightarrow$ Emotion;
Cognition $\rightarrow$ Behavior;
Emotion $\rightarrow$ Cognition;
Stimulus $\rightarrow$ Cognition;
Behavior $\rightarrow$ Cognition.
\\

\bottomrule
\end{tabular}
\end{table}

\section{Observer Model Training}
\label{app:observer_training}

This section provides the complete training procedure for the Observer Model.
The Observer converts patient profiles into directed psychological network
models and is subsequently frozen for Actor training. We decompose network
model construction into two stages: \emph{node construction} (Stage 1) and
\emph{edge construction} (Stage 2). A single Qwen3-8B model is trained
sequentially across both stages using supervised fine-tuning (SFT) followed by
Group Relative Policy Optimization (GRPO).

\subsection{Task Formulation}
\label{app:observer_task}

Let $x$ denote the free-text presenting complaints of a patient. In Stage 1,
the Observer predicts a set of psychological and external factors
\[
V=\{v_1,\ldots,v_n\}.
\]
In Stage 2, conditioned on $x$ and the predicted node set $V$, the Observer
constructs a directed network model
\[
G=(V,E),
\]
where each edge $(v_i\rightarrow v_j)\in E$ indicates that $v_i$ causes,
triggers, worsens, or maintains $v_j$.

The Observer outputs a free-form reasoning trace followed by a structured
graph:
\[
\pi_\theta(x,V)
\rightarrow
\texttt{<think>}z\texttt{</think>}
\;
\texttt{<GRAPH>}\cdots\texttt{</GRAPH>},
\]
where $z$ denotes the reasoning trace. Stage 1 outputs nodes and Stage 2
outputs directed relationships among the provided nodes.

\subsection{Training Data}
\label{app:observer_data}

We construct Observer training data from psychotherapy case reports. GPT-5 is
used to generate reference nodes, reasoning traces, and graph-construction
responses. We additionally paraphrase the presenting complaints to increase
linguistic variation while preserving the underlying case information. This
results in $5{,}599$ Stage-1 SFT examples, $2{,}138$ Stage-2 SFT examples,
and $5{,}089$ prompts for GRPO training.

\subsection{Supervised Fine-Tuning}
\label{app:observer_sft}

We first fine-tune Qwen3-8B with QLoRA to learn structured network model
construction. Each SFT example contains a system prompt specifying the graph
construction task, patient information, and a teacher-generated response
containing a reasoning trace followed by the target graph.

Training proceeds sequentially. We first perform Stage-1 SFT for node
construction. After Stage-1 GRPO, the resulting adapter is merged into the
base model. We then perform Stage-2 SFT for edge construction on top of the
Stage-1 model, followed by Stage-2 GRPO.

\subsection{Stage-1 GRPO: Node Construction}
\label{app:observer_grpo_s1}

After Stage-1 SFT, we optimize node construction using GRPO. For each patient
profile, the policy samples eight candidate outputs. The Stage-1 reward is
\begin{equation}
R_{\mathrm{S1}}
=
0.4R_{\mathrm{format}}
+
0.6R_{\mathrm{recall}},
\end{equation}
where $R_{\mathrm{format}}$ rewards valid structured output and
$R_{\mathrm{recall}}$ measures recovery of the reference nodes.

A predicted node is matched to a reference node when their MiniLM cosine
similarity is at least $0.75$. Node recall is computed as the proportion of
reference nodes successfully matched and is mapped to $[-1,1]$. Stage-1 GRPO
is performed for six rounds of 100 optimization steps, yielding 600 cumulative
GRPO steps.

\subsection{Stage-2 GRPO: Edge Construction}
\label{app:observer_reward}

Stage 2 optimizes the causal relationships among the provided nodes. For each
training example, the policy samples four candidate network models. The reward
is
\begin{equation}
R_{\mathrm{S2}}
=
0.1R_{\mathrm{format}}
+
0.6R_{\mathrm{precision}}
+
0.3R_{\mathrm{coverage}}
-
R_{\mathrm{size}}.
\label{eq:observer_s2_reward}
\end{equation}

\paragraph{Edge Precision.}
For every proposed edge $e\in E$, an independent GPT-5-mini judge assigns a
plausibility score $J(x,e)\in[0,1]$ based on whether the relationship is
supported by the presenting complaints. Valid judge scores are mapped to
$[-1,1]$:
\begin{equation}
s_e = 2J(x,e)-1.
\end{equation}
Edges containing nodes outside the provided node set, self-loops, and
duplicate edges receive a score of $-1$ without a judge call. Edge precision
is the mean plausibility of the proposed edges:
\begin{equation}
R_{\mathrm{precision}}
=
\frac{1}{|E|}
\sum_{e\in E}s_e.
\label{eq:observer_precision}
\end{equation}

\paragraph{Node Coverage.}
Optimizing edge precision alone could favor a graph containing only a small
number of high-confidence relationships. We therefore additionally reward
coverage of the provided nodes. Let
\[
E^{+}=\{e\in E:J(x,e)\geq0.5\}
\]
denote the supported edges, and let $V^{+}$ contain all provided nodes
incident to at least one edge in $E^{+}$. We define
\begin{equation}
R_{\mathrm{coverage}}
=
\frac{|V^{+}|}{|V|}.
\label{eq:observer_coverage}
\end{equation}

\paragraph{Size Penalty.}
To discourage unnecessarily dense network models, we penalize edges beyond
10:
\begin{equation}
R_{\mathrm{size}}
=
\min\left(
0.05\max(0,|E|-10),
0.3
\right).
\label{eq:observer_size}
\end{equation}

Together, these terms encourage graphs that contain plausible relationships
while covering the provided psychological factors without becoming
unnecessarily dense. Unlike within-graph normalization, the edge-plausibility
scores in \Cref{eq:observer_precision} are averaged directly and therefore
preserve differences in plausibility across candidate network models.

\subsection{Graph Plausibility Judge}
\label{app:observer_judge}

We implement the edge-plausibility judge using Azure-hosted GPT-5-mini. Given
the patient's presenting complaints and a set of proposed relationships, the
judge evaluates each relationship independently and assigns a score from 0 to
1, where higher values indicate stronger support from the patient profile.
Judge scores are cached for each patient--edge pair so that candidate
generations for the same prompt are evaluated using consistent scores.

The judge prompt is shown below.

\begin{quote}
\small
\textbf{System Prompt:}

You are a professional mental health expert.

\vspace{0.5em}

\textbf{User Prompt:}

This is the presenting complaints of the mental health patient:

\texttt{\{complaints\}}

Below are several proposed symptom relationships. For each relationship,

\begin{itemize}
    \item Evaluate whether the relationship is clinically plausible based on
    the presenting complaints.
    \item Assign a score between 0 and 1.
    \item 1 indicates that the relationship is clearly supported.
    \item 0 indicates that the relationship is clearly unsupported.
    \item 0.5 indicates uncertain or weak evidence.
\end{itemize}

Return your answer in \textbf{strict JSON} format:

\begin{verbatim}
[
  {"edge_id":0,"score":0.85},
  {"edge_id":1,"score":0.20}
]
\end{verbatim}

Symptom Relationships:

0. \texttt{\{edge\_0\}}

1. \texttt{\{edge\_1\}}

...
\end{quote}

Unless otherwise specified, the judge uses a sampling temperature of 1.0,
with up to eight concurrent requests and five retries.

\subsection{GRPO Optimization}
\label{app:observer_grpo}

For both stages, GRPO samples a group of candidate outputs for each training
example, evaluates them with the corresponding stage-specific reward, and
computes group-relative advantages. The policy is then optimized relative to
a reference policy. The overall procedure is summarized in
Algorithm~\ref{alg:observer_grpo}.

\begin{algorithm}[t]
\caption{GRPO Training for the Observer Model}
\label{alg:observer_grpo}
\begin{algorithmic}[1]
\REQUIRE SFT-initialized policy $\pi_\theta$, reference policy
$\pi_{\mathrm{ref}}$, training dataset $\mathcal{D}$, stage-specific reward
function $R$, group size $G$
\ENSURE Optimized policy $\pi_\theta$

\FOR{each training step}
    \STATE Sample a mini-batch from $\mathcal{D}$
    \FOR{each training example $x$}
        \STATE Sample $G$ candidate outputs $y_1,\ldots,y_G$
        \STATE Compute rewards $R(x,y_1),\ldots,R(x,y_G)$
        \STATE Compute group-relative advantages
    \ENDFOR
    \STATE Update $\pi_\theta$ relative to $\pi_{\mathrm{ref}}$
\ENDFOR
\end{algorithmic}
\end{algorithm}

Stage-1 GRPO uses eight generations per prompt and is trained for six
100-step rounds. Stage-2 GRPO uses four generations per prompt and is trained
for eight 100-step rounds. Each round resumes the LoRA adapter from the
previous round. The final Stage-2 adapter is merged with the preceding model
to obtain the frozen Observer used during Actor training.

\subsection{Implementation Details}
\label{app:observer_implementation}

We use Qwen3-8B as the base model and train all stages on four GPUs using
QLoRA with 4-bit NF4 quantization and bf16 computation. SFT is implemented
with \texttt{SFTTrainer} and GRPO with \texttt{GRPOTrainer}. The main
hyperparameters are summarized in \Cref{tab:observer_training}.

\begin{table*}[t]
\centering
\small
\caption{Training configuration of the Observer Model.}
\label{tab:observer_training}
\begin{tabular}{lccc}
\toprule
& \textbf{SFT} & \textbf{Stage-1 GRPO} & \textbf{Stage-2 GRPO} \\
\midrule
Base model
& Qwen3-8B
& Stage-1 SFT
& Stage-2 SFT \\

LoRA rank $r$
& 64 & 16 & 32 \\

LoRA $\alpha$
& 16 & 32 & 64 \\

Learning rate
& $1\times10^{-4}$
& $5\times10^{-6}$
& $1\times10^{-5}$ \\

GRPO generations
& -- & 8 & 4 \\

Training length
& 3 epochs
& $100\times6$ steps
& $100\times8$ steps \\

Max. prompt length
& -- & 2048 & 2048 \\

Max. completion length
& -- & 4096 & 4096 \\

Quantization
& 4-bit NF4
& 4-bit NF4
& 4-bit NF4 \\

Computation
& bf16
& bf16
& bf16 \\
\bottomrule
\end{tabular}
\end{table*}

\section{Actor Model Training}
\label{app:actor_training}

This section describes how the Actor Model learns to generate realistic patient behavior conditioned on psychological network models and their associated masks. We detail the masking strategy, conversation simulation, graph-based preference construction, Direct Preference Optimization (DPO), prompts, and implementation settings.

\subsection{Mask Strategy and Data Augmentation}
\label{app:mask}

The Actor Model is conditioned on a complete network model together with a mask that controls the initial disclosure and activation of psychological relationships. Given the original network model $N_0$ constructed from a patient profile, we define a conditioning representation $\tilde{N}_i=(N_0,M_i)$, where $M_i$ identifies relationships that are initially hidden but remain part of the underlying network model.

Masking serves two purposes: (1) modeling patients' incomplete awareness or disclosure of their psychological mechanisms, and (2) generating diverse behavioral trajectories from the same patient profile for data augmentation.

\paragraph{Node typing.}
Before masking, every node in $N_0$ is assigned one of five types: \emph{Behavior}, \emph{Emotion}, \emph{Cognition}, \emph{Physiological Sensation}, and \emph{Stimulus} (an external event), using an LLM prompted with the definitions in \Cref{app:node_types}. Each directed edge is then labeled by its transition type, i.e., \texttt{(parent type $\rightarrow$ child type)}.

\paragraph{Expert-defined mask patterns.}
Together with clinical experts, we define six disorder-specific mask patterns (\Cref{tab:mask_patterns}). Each pattern specifies edge-type transitions eligible for masking, reflecting psychological relationships that patients with the corresponding disorder may not initially recognize or disclose.

\paragraph{Masking procedure.}
Given $N_0$, we (1) classify the patient into one of six disorder categories, (2) assign node types and edge transition types, and (3) randomly select a fraction of pattern-eligible edges to mark as initially hidden. Importantly, these relationships are not removed from $N_0$. Instead, the Actor receives the complete network model together with masking instructions distinguishing initially visible relationships from hidden ones. During conversation, contextual triggers may activate previously hidden relationships, allowing them to emerge naturally in the simulated patient's behavior. We use six masking levels, $p_1,\ldots,p_6 = \{1\%, \ldots, 16\%\}$ of all relationships, to construct distinct conditioning representations from each network model (\Cref{sec:mask}).

\paragraph{Worked example.}
For the anxiety patient \emph{Victoria}, the anxiety mask pattern identifies \emph{Emotion$\rightarrow$Cognition} and \emph{Cognition$\rightarrow$Behavior} transitions as eligible for masking. At masking level $p_2$, two edges are marked as initially hidden. These relationships remain part of Victoria's underlying network model but are not initially disclosed. As the conversation unfolds, relevant triggers may activate these relationships, allowing different behavioral trajectories to emerge from the same psychological structure.

\subsection{Conversation Simulation and Supervised Fine-Tuning}
\label{app:actor_sft}

\paragraph{Conversation simulation.}
To construct supervised fine-tuning data, a Qwen3-30B patient simulator, conditioned on $\tilde{N}_i=(N_0,M_i)$, role-plays a vulnerable user while interacting with a partner therapist model (GPT-4o via Azure OpenAI), yielding $423$ patient--therapist transcripts. The patient simulator receives the complete network model and instructions specifying which relationships are initially hidden. Initially visible relationships guide the patient's psychological responses, while hidden relationships may become activated through contextual triggers during the conversation.

The patient simulator is instructed to generate responses in the format
\[
\texttt{<state>...</state>}
\quad
\texttt{<word>...</word>},
\]
where the internal state records newly activated psychological states and the utterance expresses the corresponding first-person experience. The Actor is instructed not to explicitly mention the network model or its mechanisms in the natural-language response.

\paragraph{Supervised Fine-Tuning.}
The simulated conversations are converted into instruction-following training examples. Given $\tilde{N}_i$ and the dialogue history, the Actor Model learns to generate structured patient responses that express the underlying psychological mechanisms while maintaining natural conversational behavior. These supervised demonstrations initialize the Actor for subsequent preference optimization. During preference optimization (\Cref{app:actor_dpo}), conversations are instead generated by the SFT-initialized Qwen3-8B Actor.

\subsection{Graph Similarity Reward}
\label{app:actor_reward}

\paragraph{Network model reconstruction.}
After each simulated conversation, the frozen Observer Model reconstructs a network model $N_{i,j}$ from the simulated patient's observable utterances using the extraction prompt described in \Cref{app:prompts}. The Observer is constrained to predict relationships among the provided candidate nodes, ensuring that the reconstructed and conditioning network models share the same node vocabulary.

\paragraph{Graph similarity.}
We evaluate how faithfully a simulated conversation expresses the conditioning psychological mechanisms using directed-edge $F_1$. Let $E(N_{i,j})$ denote the reconstructed directed edges and $E_{\mathrm{target},i}$ the target edge set associated with the conditioning representation $\tilde{N}_i$. A predicted edge is considered correct only when both endpoints and its direction match a target edge.

We compute precision and recall as
\begin{equation}
P=\frac{\mathrm{TP}}{|E(N_{i,j})|},\qquad
R=\frac{\mathrm{TP}}{|E_{\mathrm{target},i}|},
\end{equation}
where
\[
\mathrm{TP}
=
|E(N_{i,j})\cap E_{\mathrm{target},i}|.
\]

The graph reconstruction score is
\begin{equation}
\label{eq:graph_similarity}
s_{i,j}
=
\operatorname{sim}(N_{i,j},\tilde{N}_i)
=
\frac{2PR}{P+R}.
\end{equation}

We use this score to rank simulated conversations and construct preference pairs for subsequent DPO training. A reference implementation is provided in \Cref{app:f1code}.

\subsection{Preference Construction and Direct Preference Optimization}
\label{app:actor_dpo}

\paragraph{Preference construction.}

For each conditioning representation $\tilde{N}_i=(N_0,M_i)$,
the Actor generates $K=5$ candidate conversations under diverse
decoding configurations. After each conversation $C_{i,j}$,
the frozen Observer reconstructs a network model $N_{i,j}$
and computes its graph reconstruction score
$s_{i,j}=\operatorname{sim}(N_{i,j},\tilde{N}_i)$,
as defined in \Cref{eq:graph_similarity}.

For each conditioning representation, we select the
highest-scoring and lowest-scoring conversations as the
preferred and rejected samples, respectively, provided
that their score difference exceeds a threshold $\delta$.
This procedure yields approximately 6,900 preference pairs
for subsequent optimization.

\paragraph{Direct Preference Optimization.}

We optimize the Actor using Direct Preference Optimization
(DPO). Given the constructed preference pairs, DPO encourages
the Actor to assign higher relative likelihood to conversations
whose reconstructed network models better match the target
psychological relationships.

Let $x$ denote the conditioning representation together with
the dialogue context, and let $y_w$ and $y_l$ denote the
preferred and rejected conversation continuations.
The objective is

\begin{equation}
\begin{aligned}
\mathcal{L}_{\mathrm{DPO}}
= -\mathbb{E}_{(x,y_w,y_l)\sim\mathcal{P}}
\Bigg[
\log \sigma\Bigg(
&\beta \log
\frac{\pi_\theta(y_w\mid x)}
{\pi_{\mathrm{ref}}(y_w\mid x)}
\\[-2pt]
&-\beta \log
\frac{\pi_\theta(y_l\mid x)}
{\pi_{\mathrm{ref}}(y_l\mid x)}
\Bigg)
\Bigg].
\end{aligned}
\end{equation}

Here, $\pi_\theta$ is the optimized Actor,
$\pi_{\mathrm{ref}}$ is the frozen SFT reference policy,
and $\beta$ controls the strength of the preference
optimization relative to the reference policy.

This objective encourages the Actor to express the
conditioning psychological mechanisms through conversational
behavior while remaining close to the SFT policy.
The Actor is additionally instructed not to explicitly
mention the network model or its mechanisms, discouraging
solutions that verbalize the graph rather than expressing
it through natural first-person experiences.

\subsection{Training Configuration}
\label{app:actor_config}

\Cref{tab:actorhp} summarizes the training configuration
of the Actor Model. We initialize the Actor through
supervised fine-tuning (SFT) using QLoRA with 4-bit NF4
quantization and bf16 precision. The resulting SFT model
is subsequently optimized with DPO, using the frozen
SFT policy as the reference. During DPO, only the LoRA
parameters are updated, while the backbone remains frozen.
Unless otherwise specified, all experiments use the
hyperparameters in \Cref{tab:actorhp}.

\begin{table}[t]
\centering
\small
\begin{tabular}{lcc}
\toprule
 & \textbf{SFT} & \textbf{DPO} \\
\midrule
Base model & Qwen3-8B & Qwen3-8B + SFT adapter \\
Quantization & 4-bit NF4, bf16 & 4-bit NF4, bf16 \\
LoRA rank $r$ & 64 & 16 \\
LoRA $\alpha$ & 16 & 32 \\
LoRA dropout & 0.05 & 0.05 \\
LoRA targets & \multicolumn{2}{c}{q,k,v,o,gate,up,down proj} \\
Learning rate & $1\times10^{-4}$ & $2\times10^{-6}$ \\
Optimizer & paged AdamW-32bit & AdamW \\
Effective batch size & 16 ($4\times4$) & 16 ($2\times8$) \\
Epochs & 5 & 2 \\
Maximum sequence length & 2048 & 4096 (prompt 2048) \\
DPO $\beta$ & --- & 0.1 \\
Reference policy & --- & Frozen SFT policy \\
Precision & bf16 & bf16 \\
\bottomrule
\end{tabular}
\caption{Training configuration of the Actor Model.}
\label{tab:actorhp}
\end{table}

\subsection{Prompts}
\label{app:prompts}

\paragraph{Node-typing prompt (mask strategy).}
\begin{quote}\scriptsize\ttfamily
Please classify each symptom into 5 types: Behavior, Emotion, Physiological
Sensation, Cognition and Stimulus (i.e., an event). This is the Symptom List:
\{symptoms\}. [Definitions: Behaviors are discrete actions people take, voluntary
or involuntary (involuntary bodily events such as sweating/heartbeat are
Physiological Sensations). Emotions are subjective feelings named by an emotion
word or valence/arousal. Cognitions are thoughts, appraisals, interpretations,
beliefs, or predictions about oneself, others, the world, or the future---often
phrased ``I feel like\ldots'' without an emotion word. A Stimulus is an external
event perceived through the senses. A Physiological Sensation is an internal
interoceptive experience produced by the body's physiological systems.] Give each
symptom a type. Enclose the explanation within <exp>...</exp> and the type list
within <type>[Behavior, Emotion, Cognition, ...]</type>.
\end{quote}

\paragraph{Disorder-classification prompt (mask strategy).}
\begin{quote}\scriptsize\ttfamily
Below are the patient's symptoms:\\
\{symptoms\}\\
Please classify the patient into one of the following categories:\\
\{depression, anxiety, substance use, trauma, mania, psychosis\}\\
Only output the category name. Do not provide any explanation.
\end{quote}

\paragraph{Patient roleplay prompt (Actor generation).}
\begin{quote}\scriptsize\ttfamily
You are role-playing a mental health patient. Your internal experience must
strictly follow a directed symptom activation system.\\[2pt]
Definition: If a source state becomes active due to a concrete real-life event,
the linked target state becomes more likely next. Hidden or subconscious states
may emerge naturally when triggered, but are initially unrecognized by the
patient. Do NOT mention the graph or mechanisms.\\[2pt]
ALLOWED STATES (no additions, no substitutions): \{nodes grouped by
Cognition/Emotion/Behavior/Stimulus; hidden nodes annotated ``(initially
subconscious, may emerge during conversation)''\}\\[2pt]
MANDATORY DIRECTED ACTIVATION RULES: \{numbered visible edges; hidden edges
annotated ``(hidden activation, initially subconscious)''\}. Constraints: do not
reverse edges; do not skip intermediate links; do not invent additional edges.\\[2pt]
EXPRESSION REQUIREMENTS (STRICT): speak in first person; do not repeat previous
statements; no clinical labels; no abstract summaries; every internal shift must
be tied to a specific real-life event; narrative structure
event $\rightarrow$ interpretation $\rightarrow$ emotional/physiological change
$\rightarrow$ behavior. Every response must be wrapped as:\\
<state>Your new state during the conversation.</state>\\
<word>Your first-person narrative responding to the therapist.</word>\\[2pt]
[If suicidality is present:] SAFETY RULE --- suicidal thoughts must remain passive
and vague, with no intent, planning, or method. Remain fully consistent with this
activation system in every response.
\end{quote}

\paragraph{Therapist prompt (partner model, \texttt{GPT-4o}).}
\begin{quote}\scriptsize\ttfamily
You are a licensed mental health therapist (e.g., clinical psychologist or
licensed counselor) trained in evidence-based practice. You must always speak as
the therapist only. Guidelines: maintain a professional, ethical, empathetic tone;
use CBT-style Socratic questioning; do NOT diagnose; do NOT role-play the patient;
keep responses under 200 words; end with a reflective or exploratory question.
\end{quote}

\paragraph{Graph-reconstruction prompt (Observer).}
\begin{quote}\scriptsize\ttfamily
You are a clinical information extraction model. Given the Symptoms and External
Factor List and the patient's Presenting Complaints, construct a directed symptom
network. Rules: use only symptoms or external factors from the provided list;
create a link A $\rightarrow$ B only if the text supports that A causes, triggers,
worsens, or maintains B; the direction must reflect cause $\rightarrow$ effect; do
not infer from general medical knowledge or co-occurrence; do not create links
involving symptoms not explicitly mentioned; reasoning $\le$ 500 words. Put
reasoning between <think> and </think>; output only valid JSON between <GRAPH> and
</GRAPH> in the form \{"links":[\{"from":...,"to":...\}]\}.
\end{quote}

\subsection{Graph Similarity Implementation}
\label{app:f1code}

Graph similarity provides the scoring signal used to construct
preference pairs for Actor optimization. We compute directed-edge
$F_1$ between the reconstructed network model $N_{i,j}$
and a target edge set $E_{\mathrm{target},i}$ derived from
the conditioning representation $\tilde{N}_i=(N_0,M_i)$.
An edge is considered correct only when both endpoints
and its direction match a target relationship.

The implementation normalizes node names by lowercasing
and trimming whitespace, converts each graph into a set
of directed edges, and computes precision, recall,
and their harmonic mean. This corresponds to
\Cref{eq:graph_similarity}.

\begin{verbatim}
def graph_similarity(target_edges, reconstructed_edges):
    """Directed-edge F1 for graph reconstruction."""

    def normalize(edge):
        return (
            edge["from"].strip().lower(),
            edge["to"].strip().lower(),
        )

    target = {normalize(e) for e in target_edges}
    reconstructed = {normalize(e) for e in reconstructed_edges}

    if not target and not reconstructed:
        return 1.0
    if not target or not reconstructed:
        return 0.0

    tp = len(target & reconstructed)

    precision = tp / len(reconstructed)
    recall = tp / len(target)

    if precision + recall == 0:
        return 0.0

    return 2 * precision * recall / (precision + recall)
\end{verbatim}

\section{Profile Expansion Evaluation Details}
\label{app:experimental_protocols}

This section provides implementation details for the profile expansion
evaluation, including the complete definitions of our evaluation metrics,
the structured interview protocol, and the topic--attribute schema used for
Behavior Diversity.

\subsection{Evaluation Metrics}
\label{app:evaluation_metrics}

We provide the complete definitions and implementation details for
Simulation Diversity, Behavior Diversity, and Profile Alignment below.

\subsubsection{Simulation Diversity}
\label{app:simulation_diversity}

Simulation Diversity measures semantic variation across independently
generated responses to the same interview topic. For each topic, we first
filter responses that are too short or lexically sparse for reliable semantic
comparison. A response is considered \emph{informative} if it satisfies
minimum thresholds for both response length and the number of distinct tokens.

For a given model, profile, and topic $t$, let
\[
\mathcal{I}_t=\{\mathbf{e}_i\}_{i=1}^{n_t}
\]
denote the $L_2$-normalized embeddings of informative responses across
simulation runs. Topic-level diversity is the average distance from each
response to its nearest semantic neighbor:
\[
D_t
=
\frac{1}{n_t}
\sum_{i=1}^{n_t}
\min_{\substack{1\le j\le n_t\\ j\neq i}}
\left[
1-\cos(\mathbf{e}_i,\mathbf{e}_j)
\right],
\qquad n_t\ge2.
\]

Let $\mathcal{T}=\{t:n_t\ge2\}$ denote topics with at least two informative
responses. We average across these topics and adjust for response coverage:
\[
D
=
\left(
\frac{1}{|\mathcal{T}|}
\sum_{t\in\mathcal{T}}D_t
\right)
\left(
\frac{N_{\mathrm{info}}}{N_{\mathrm{total}}}
\right),
\]
where $N_{\mathrm{info}}$ and $N_{\mathrm{total}}$ are the numbers of
informative and total responses, respectively. This coverage term prevents a
model from receiving a high diversity score based on only a small subset of
its generations. We set $D=0$ when $\mathcal{T}=\varnothing$ and report the
mean score across profiles.

\subsubsection{Behavior Diversity}
\label{app:behavior_diversity}

Behavior Diversity measures variation in substantive user characteristics
across repeated simulations rather than variation in surface wording.
We use a predefined schema containing 141 attributes across 14 interview
topics, detailed in \Cref{app:behavior_diversity_schema}.

For each simulation run $r$, topic $t$, and attribute $a$, let
$V_{r,t,a}$ denote the normalized set of extracted values. Variation between
two runs is measured using Jaccard distance:
\begin{equation}
d_{t,a}(r,s)
=
1-
\frac{
|V_{r,t,a}\cap V_{s,t,a}|
}{
|V_{r,t,a}\cup V_{s,t,a}|
}.
\end{equation}

Pairs for which both value sets are empty are excluded because they provide no
evidence about behavioral variation. All other pairs, including pairs in which
only one run contains an extracted value, are retained. Let
\begin{equation}
\mathcal{P}_{t,a}
=
\left\{
(r,s):r<s,\;
V_{r,t,a}\cup V_{s,t,a}\neq\varnothing
\right\}
\end{equation}
denote the set of valid run pairs for attribute $a$ under topic $t$. We compute
attribute-level diversity as
\begin{equation}
D_{t,a}
=
\frac{1}{|\mathcal{P}_{t,a}|}
\sum_{(r,s)\in\mathcal{P}_{t,a}}
d_{t,a}(r,s).
\end{equation}

Let
\begin{equation}
\mathcal{A}_t^{*}
=
\left\{
a:|\mathcal{P}_{t,a}|>0
\right\}
\end{equation}
denote the set of scorable attributes for topic $t$. Topic-level Behavior
Diversity is
\begin{equation}
D_t
=
\frac{1}{|\mathcal{A}_t^{*}|}
\sum_{a\in\mathcal{A}_t^{*}}
D_{t,a}.
\end{equation}

Finally, let
\begin{equation}
\mathcal{T}^{*}
=
\left\{
t:|\mathcal{A}_t^{*}|>0
\right\}.
\end{equation}
Profile-level Behavior Diversity is
\begin{equation}
D_{\mathrm{behavior}}
=
\frac{1}{|\mathcal{T}^{*}|}
\sum_{t\in\mathcal{T}^{*}}D_t.
\end{equation}

The final model-level score is the mean profile-level Behavior Diversity
across evaluated profiles. All scorable attributes and topics receive equal
weight. We report the raw Jaccard-based diversity score without
topic-specific offsets, clipping transformations, or nonlinear rescaling.

We additionally report attribute coverage separately from diversity, defined
as the proportion of predefined attributes for which at least one value is
extracted across the repeated simulations. Separating coverage from diversity
prevents differences in the amount of expressed information from being
implicitly incorporated into the diversity score.

\subsubsection{Profile Alignment}
\label{app:profile_alignment}

Profile Alignment measures whether a simulated user remains faithful to the
assigned profile throughout the conversation. We use
GPT-5~\cite{openai2025gpt5} as an LLM judge to evaluate five aspects:
\emph{background}, \emph{presenting problem}, \emph{emotion},
\emph{interpersonal style}, and \emph{symptoms}.
Each aspect is rated on an anchored five-point scale.

For a conversation, let $s_k\in\{1,\ldots,5\}$ denote the rating for aspect
$k$, and let
\[
\mathcal{K}_{\mathrm{valid}}
=
\{k\in\mathcal{K}:s_k\in\{1,\ldots,5\}\}.
\]
We normalize ratings to $[0,1]$ and compute
\[
S_{\mathrm{align}}
=
\begin{cases}
\displaystyle
\frac{1}{|\mathcal{K}_{\mathrm{valid}}|}
\sum_{k\in\mathcal{K}_{\mathrm{valid}}}
\frac{s_k-1}{4},
&
|\mathcal{K}_{\mathrm{valid}}|>0,\\[8pt]
0,
&
|\mathcal{K}_{\mathrm{valid}}|=0.
\end{cases}
\]
Higher scores indicate greater consistency with the assigned profile. We
report the mean across conversations and profiles.

\subsection{Profile Expansion Interview Agenda}
\label{app:profile_expansion_agenda}

To ensure that all simulated patients are interviewed under a consistent and comprehensive protocol, we use a structured clinical interview agenda. The agenda contains nine sessions covering the presenting problem, major symptoms, history of the problem, functional impact, coping, social support, treatment history, risk assessment, and treatment goals.

For each topic, we provide the session name, assessment goal, opening question, and optional follow-up questions. Interviewers begin with the opening question and use the follow-up questions as needed to clarify or deepen the patient's response. The questions are intended to guide the interview rather than impose a fixed script.

\subsubsection{Session 1: Presenting Problem / Chief Complaint}
\label{app:agenda_presenting_problem}

\paragraph{Topic.}
Presenting problem / chief complaint.

\paragraph{Goal.}
Understand why the patient is seeking help now, how they describe the main concern in their own words, and what they hope to receive from the interview or treatment.

\paragraph{Opening Question.}
\begin{quote}
What brings you in today?
\end{quote}

\paragraph{Follow-up Questions.}
\begin{itemize}
    \item What are you hoping we can help you with?
    \item How would you describe the problem?
    \item What feels most important for me to understand about what you are going through?
\end{itemize}

\subsubsection{Session 2: Major Symptoms}
\label{app:agenda_major_symptoms}

\paragraph{Topic A: Emotions.}

\paragraph{Goal.}
Understand the patient's emotional experience.

\paragraph{Opening Question.}
\begin{quote}
What emotions have you been feeling most often?
\end{quote}

\paragraph{Follow-up Questions.}
\begin{itemize}
    \item What feels most distressing emotionally?
    \item How intense do these feelings get?
    \item How often do these emotions come up?
\end{itemize}

\paragraph{Topic B: Behaviors.}

\paragraph{Goal.}
Understand how the problem appears in the patient's behavior.

\paragraph{Opening Question.}
\begin{quote}
What do you do when the problem gets worse?
\end{quote}

\paragraph{Follow-up Questions.}
\begin{itemize}
    \item Have you changed how you act or what you do day to day?
    \item Are there things you avoid because of the problem?
    \item Have other people noticed changes in your behavior?
\end{itemize}

\paragraph{Topic C: Cognitions.}

\paragraph{Goal.}
Understand the patient's thoughts, beliefs, worries, interpretations, and cognitive patterns.

\paragraph{Opening Question.}
\begin{quote}
What thoughts usually come up when you feel distressed?
\end{quote}

\paragraph{Follow-up Questions.}
\begin{itemize}
    \item What do you worry about most?
    \item What do you tell yourself about the situation?
    \item How do you interpret what is happening to you?
\end{itemize}

\subsubsection{Session 3: History of the Problem}
\label{app:agenda_history}

\paragraph{Topic A: Onset and Timeline.}

\paragraph{Goal.}
Establish the chronology and course of the presenting problem.

\paragraph{Opening Question.}
\begin{quote}
How long has this been a concern?
\end{quote}

\paragraph{Follow-up Questions.}
\begin{itemize}
    \item When did this first begin?
    \item Is this the first time you have had these problems?
    \item If not, how many times has this happened before, and when was the first time?
    \item Has the problem changed over time?
\end{itemize}

\paragraph{Topic B: Triggers and Precipitating Events.}

\paragraph{Goal.}
Identify events, stressors, or situations that started, reactivated, or worsened the problem.

\paragraph{Opening Question.}
\begin{quote}
According to you, what caused the problem?
\end{quote}

\paragraph{Follow-up Questions.}
\begin{itemize}
    \item Were there any new events happening in your life, or anything stressful?
    \item What events or situations make the problem worse?
    \item Are there situations where the problem feels better?
    \item Can you tell me more about how these events are linked to your problem?
\end{itemize}

\subsubsection{Session 4: Impact on Functioning}
\label{app:agenda_functioning}

\paragraph{Topic.}
Impact on functioning.

\paragraph{Goal.}
Assess how the problem affects the patient's daily life, including work or school, relationships, health, self-care, and everyday tasks.

\paragraph{Opening Question.}
\begin{quote}
How has this problem affected your life?
\end{quote}

\paragraph{Follow-up Questions.}
\begin{itemize}
    \item Can you do your usual work or everyday tasks?
    \item Have you noticed any changes in your relationships?
    \item How has it affected your work, school, health, or daily responsibilities?
    \item What does a typical day look like for you now?
\end{itemize}

\subsubsection{Session 5: Current Coping}
\label{app:agenda_coping}

\paragraph{Topic.}
Current coping.

\paragraph{Goal.}
Understand how the patient currently deals with the problem, what helps, what makes it worse, and whether avoidance or other maladaptive coping patterns are present.

\paragraph{Opening Question.}
\begin{quote}
How have you been coping or dealing with this?
\end{quote}

\paragraph{Follow-up Questions.}
\begin{itemize}
    \item What helps, and what makes it worse?
    \item Are there times when the problem feels better or worse?
    \item What do you usually do when the distress becomes strong?
\end{itemize}

\subsubsection{Session 6: Social Support}
\label{app:agenda_social_support}

\paragraph{Topic.}
Social support.

\paragraph{Goal.}
Assess who is important in the patient's life, the quality of available support, whether the patient feels isolated, and what others have suggested.

\paragraph{Opening Question.}
\begin{quote}
Who are the important people in your life that you see or talk to regularly?
\end{quote}

\paragraph{Follow-up Questions.}
\begin{itemize}
    \item Do you have close friends or family?
    \item Would you like to have more friends or support?
    \item What do your family, friends, or other people suggest you do?
    \item Who do you usually turn to when things are difficult?
\end{itemize}

\subsubsection{Session 7: Past Treatment / Psychiatric History}
\label{app:agenda_past_treatment}

\paragraph{Topic.}
Past treatment / psychiatric history.

\paragraph{Goal.}
Collect prior treatment experiences, psychiatric diagnoses, hospitalizations, medication history, previous episodes, family mental health history, and barriers to seeking help.

\paragraph{Opening Question.}
\begin{quote}
Have you ever received help for this or similar problems?
\end{quote}

\paragraph{Follow-up Questions.}
\begin{itemize}
    \item What kind of help did you receive, when, for how long, and what was useful or not?
    \item Have you ever received a formal mental health diagnosis?
    \item Have you ever been hospitalized for psychological problems or stressors?
    \item Have you had prior psychological or psychiatric treatment?
    \item Has anything gotten in the way of getting help?
    \item Are you aware of similar concerns among family members?
\end{itemize}

\subsubsection{Session 8: Risk Assessment}
\label{app:agenda_risk}

\paragraph{Topic A: Suicide and Self-Harm.}

\paragraph{Goal.}
Assess safety concerns related to suicide or self-harm, including frequency, plans, intent, access to means, risk factors, and protective factors.

\paragraph{Opening Question.}
\begin{quote}
Have you had thoughts that life was not worth living?
\end{quote}

\paragraph{Follow-up Questions.}
\begin{itemize}
    \item Have you thought you would be better off dead?
    \item Have you attempted to hurt yourself or end your life?
    \item How often do these thoughts occur, and when was the last time?
    \item Have you made plans to end your life, or thought about how you might go about it?
    \item Have you come close to acting on these thoughts?
    \item Do you have access to medications, guns, or other means?
    \item What keeps you going, and who do you go to when you feel this way?
\end{itemize}

\paragraph{Topic B: Harm to Others and Victimization.}

\paragraph{Goal.}
Assess risk of harm to others and any recent abuse or victimization.

\paragraph{Opening Question.}
\begin{quote}
What does it look like when you get really angry or upset?
\end{quote}

\paragraph{Follow-up Questions.}
\begin{itemize}
    \item Have you ever had thoughts, made statements, or attempted to hurt others?
    \item Have you recently been physically hurt or threatened by someone else?
\end{itemize}

\paragraph{Topic C: Substance Use.}

\paragraph{Goal.}
Assess substance use patterns and their impact on the patient's life.

\paragraph{Opening Question.}
\begin{quote}
Do you use alcohol, drugs, tobacco, or prescription medications?
\end{quote}

\paragraph{Follow-up Questions.}
\begin{itemize}
    \item How much do you use, and how often?
    \item How long has this been going on?
    \item Has substance use affected your work, school, relationships, or health?
    \item Has anyone expressed concern about your use, or have there been any legal problems?
\end{itemize}

\subsubsection{Session 9: Treatment Goals and Hope}
\label{app:agenda_treatment_goal}

\paragraph{Topic.}
Treatment goal / hope.

\paragraph{Goal.}
Understand the patient's goals for treatment, expectations for therapy, desired changes, strengths, sources of resilience, and any remaining concerns.

\paragraph{Opening Question.}
\begin{quote}
What are your goals for therapy?
\end{quote}

\paragraph{Follow-up Questions.}
\begin{itemize}
    \item What do you picture therapy to look like?
    \item What are you hoping will change?
    \item What do you think are your greatest strengths?
    \item Is there anything else we may have missed or have not covered today?
\end{itemize}

\subsection{Topic--Attribute Schema for Behavior Diversity}
\label{app:behavior_diversity_schema}

To compute behavior diversity, we organize the information elicited during
the clinical interview using a structured topic--attribute schema. Each
interview topic is associated with a predefined set of attributes describing
the patient's presentation, experiences, behaviors, history, functioning,
risk factors, and treatment goals.

Given a simulated interview, we extract the values expressed for each
attribute. Behavior diversity is computed by comparing the extracted
attribute values across repeated simulations of the same patient profile.
This topic-controlled representation distinguishes substantive differences
in simulated patient behavior from differences arising only from wording or
conversational style.

The topics correspond to the interview agenda described in
Appendix~\ref{app:profile_expansion_agenda}. The complete schema contains
14 topics and 141 topic-specific attributes.

\paragraph{Presenting Problem / Chief Complaint.}
\textbf{Attributes:}
\texttt{chief\_complaint},
\texttt{problem\_in\_own\_words},
\texttt{main\_concern},
\texttt{problem\_framing},
\texttt{reason\_for\_seeking\_help},
\texttt{current\_urgency}, and
\texttt{expectations\_for\_help}.

\paragraph{Major Symptoms: Emotions.}
\textbf{Attributes:}
\texttt{dominant\_emotion},
\texttt{secondary\_emotions},
\texttt{emotional\_intensity},
\texttt{emotional\_frequency},
\texttt{emotional\_duration},
\texttt{mood\_changes},
\texttt{anxiety\_or\_fear},
\texttt{anger\_or\_irritability},
\texttt{sadness\_or\_hopelessness}, and
\texttt{shame\_or\_guilt}.

\paragraph{Major Symptoms: Behaviors.}
\textbf{Attributes:}
\texttt{avoidance\_behavior},
\texttt{withdrawal},
\texttt{compulsions\_or\_repetitive\_behaviors},
\texttt{changes\_in\_activity},
\texttt{sleep\_behavior\_changes},
\texttt{eating\_behavior\_changes},
\texttt{agitation\_or\_restlessness},
\texttt{functional\_behavioral\_changes}, and
\texttt{maladaptive\_behavior\_patterns}.

\paragraph{Major Symptoms: Cognitions.}
\textbf{Attributes:}
\texttt{negative\_thoughts},
\texttt{worries},
\texttt{rumination},
\texttt{self\_critical\_beliefs},
\texttt{beliefs\_about\_others},
\texttt{beliefs\_about\_the\_future},
\texttt{interpretations\_of\_events},
\texttt{thought\_constriction},
\texttt{obsessional\_thoughts}, and
\texttt{hopeless\_thoughts}.

\paragraph{History of Problem: Onset / Timeline.}
\textbf{Attributes:}
\texttt{onset\_time},
\texttt{duration},
\texttt{first\_episode\_or\_recurrent\_problem},
\texttt{number\_of\_episodes},
\texttt{course\_over\_time},
\texttt{changes\_in\_severity},
\texttt{periods\_of\_improvement},
\texttt{periods\_of\_worsening}, and
\texttt{past\_similar\_problems}.

\paragraph{History of Problem: Triggers / Precipitating Events.}
\textbf{Attributes:}
\texttt{perceived\_cause},
\texttt{precipitating\_events},
\texttt{recent\_stressors},
\texttt{trigger\_events},
\texttt{trigger\_contexts},
\texttt{trigger\_frequency},
\texttt{trigger\_specificity},
\texttt{patient\_interpretation\_of\_triggers},
\texttt{situations\_that\_worsen\_the\_problem}, and
\texttt{situations\_that\_improve\_the\_problem}.

\paragraph{Impact on Functioning.}
\textbf{Attributes:}
\texttt{work\_or\_school\_impact},
\texttt{relationship\_impact},
\texttt{everyday\_task\_impact},
\texttt{self\_care\_impact},
\texttt{health\_impact},
\texttt{social\_functioning\_impact},
\texttt{role\_functioning\_impact},
\texttt{typical\_day\_disruption}, and
\texttt{functional\_impairment\_level}.

\paragraph{Current Coping.}
\textbf{Attributes:}
\texttt{coping\_strategies},
\texttt{helpful\_coping},
\texttt{maladaptive\_coping},
\texttt{avoidance\_behavior},
\texttt{safety\_behaviors},
\texttt{emotion\_regulation\_strategies},
\texttt{problem\_solving\_attempts},
\texttt{coping\_effectiveness},
\texttt{barriers\_to\_coping},
\texttt{situations\_when\_better}, and
\texttt{situations\_when\_worse}.

\paragraph{Social Support.}
\textbf{Attributes:}
\texttt{important\_people},
\texttt{support\_sources},
\texttt{support\_quality},
\texttt{frequency\_of\_contact},
\texttt{family\_support},
\texttt{friend\_support},
\texttt{community\_support},
\texttt{isolation\_level},
\texttt{desired\_support},
\texttt{barriers\_to\_support}, and
\texttt{others\_suggestions}.

\paragraph{Past Treatment / Psychiatric History.}
\textbf{Attributes:}
\texttt{prior\_help\_seeking},
\texttt{therapy\_history},
\texttt{medication\_history},
\texttt{prior\_diagnoses},
\texttt{hospitalization\_history},
\texttt{outpatient\_treatment\_history},
\texttt{significant\_psychiatric\_episodes},
\texttt{treatment\_duration},
\texttt{treatment\_response},
\texttt{helpful\_treatment\_elements},
\texttt{unhelpful\_treatment\_elements},
\texttt{barriers\_to\_treatment}, and
\texttt{family\_mental\_health\_history}.

\paragraph{Risk Assessment: Suicide / Self-Harm.}
\textbf{Attributes:}
\texttt{passive\_suicidal\_ideation},
\texttt{active\_suicidal\_ideation},
\texttt{ideation\_intensity},
\texttt{ideation\_frequency},
\texttt{ideation\_specificity},
\texttt{ambivalence},
\texttt{prior\_attempts},
\texttt{self\_harm\_history},
\texttt{plan},
\texttt{means\_access},
\texttt{means\_lethality},
\texttt{rehearsal\_or\_preparation},
\texttt{last\_occurrence}, and
\texttt{protective\_factors}.

\paragraph{Risk Assessment: Harm to Others / Victimization.}
\textbf{Attributes:}
\texttt{anger\_control},
\texttt{violent\_ideation},
\texttt{homicidal\_ideation},
\texttt{threats\_or\_statements},
\texttt{past\_violence},
\texttt{current\_victimization},
\texttt{recent\_threats}, and
\texttt{abuse\_concerns}.

\paragraph{Risk Assessment: Substance Use.}
\textbf{Attributes:}
\texttt{alcohol\_use},
\texttt{drug\_use},
\texttt{prescription\_medication\_misuse},
\texttt{tobacco\_use},
\texttt{use\_intensity},
\texttt{use\_frequency},
\texttt{use\_duration},
\texttt{substance\_related\_impairment},
\texttt{legal\_problems\_related\_to\_use}, and
\texttt{others\_concern\_about\_use}.

\paragraph{Treatment Goal / Hope.}
\textbf{Attributes:}
\texttt{treatment\_goals},
\texttt{desired\_changes},
\texttt{therapy\_expectations},
\texttt{hopefulness},
\texttt{motivation\_for\_treatment},
\texttt{strengths},
\texttt{resilience\_factors},
\texttt{hobbies\_or\_activities},
\texttt{spiritual\_or\_religious\_support}, and
\texttt{additional\_concerns}.

\section{Additional Analyses}
\label{app:additional_analysis}

We provide additional analyses of metric sensitivity, stage-level performance,
and construct validity. These analyses complement the aggregate results in the
main paper by examining how the evaluation metrics behave across different
numbers of simulation runs, interview stages, and patient profiles.

\subsection{Effect of Run Count}
\label{app:run_count}

\begin{figure*}[t]
    \centering

    \begin{minipage}[t]{0.49\textwidth}
        \centering
        \includegraphics[width=\linewidth]
        {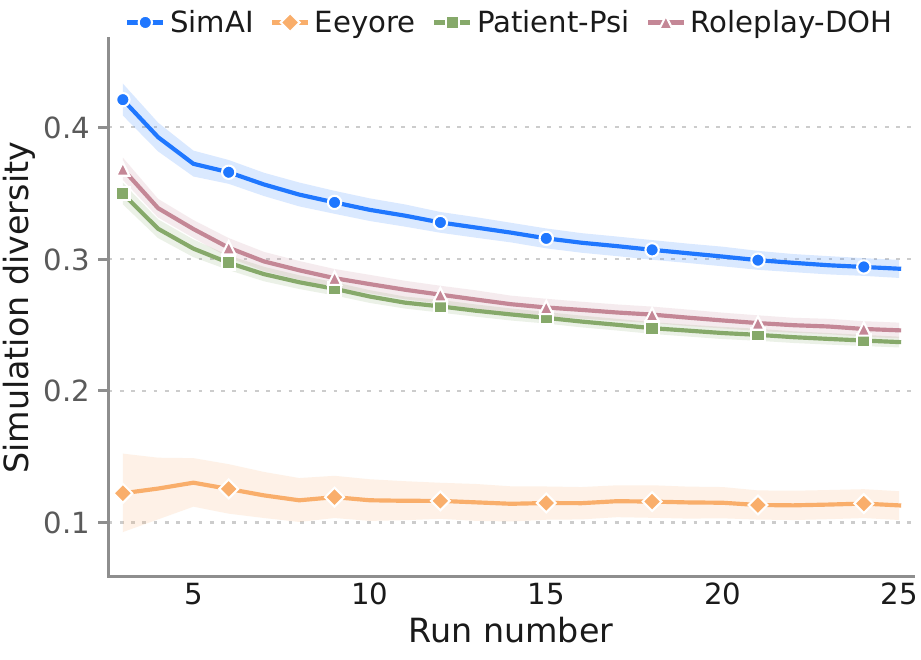}
    \end{minipage}
    \hfill
    \begin{minipage}[t]{0.49\textwidth}
        \centering
        \includegraphics[width=\linewidth]
        {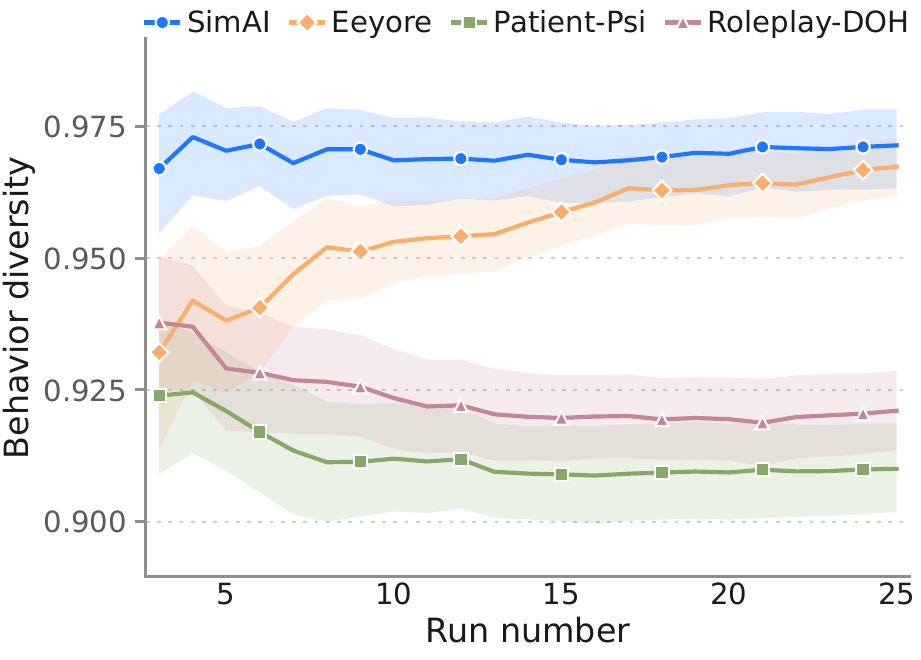}
    \end{minipage}

    \caption{
        \textbf{Effect of run count on diversity metrics.}
        Lines show mean scores across profiles, and shaded regions indicate
        95\% bootstrap confidence intervals.
        \textbf{(a)} Simulation Diversity decreases and gradually stabilizes
        as more runs are included, while \model consistently maintains the
        highest diversity.
        \textbf{(b)} Behavior Diversity is comparatively stable across run
        counts, with \model maintaining strong diversity throughout.
    }
    \label{fig:diversity_runs_line}
\end{figure*}

We examine the sensitivity of the diversity metrics by varying the number of
simulations per profile from $3$ to $25$ across the same $45$ test profiles.
As shown in \Cref{fig:diversity_runs_line}, Simulation Diversity decreases and
becomes more stable as the number of runs increases. This behavior is expected
from its nearest-neighbor formulation: as additional samples are included,
each response has more opportunities to find a semantically similar neighbor,
which tends to reduce the nearest-neighbor distance. Absolute Simulation
Diversity scores should therefore be compared at the same run count.

Behavior Diversity is comparatively stable across run counts. Because the
metric averages pairwise Jaccard distances over extracted behavioral
attributes, increasing the number of runs primarily provides additional
observations of the behavioral realizations associated with each profile.
The set of scorable attributes may also change as additional attribute values
are observed. Despite these metric-specific effects, the overall model
patterns remain relatively stable across the evaluated run counts.

\subsection{Stage- and Aspect-Level Analysis}
\label{app:stage_analysis}

\begin{figure*}[t]
    \centering
    \includegraphics[width=\textwidth]{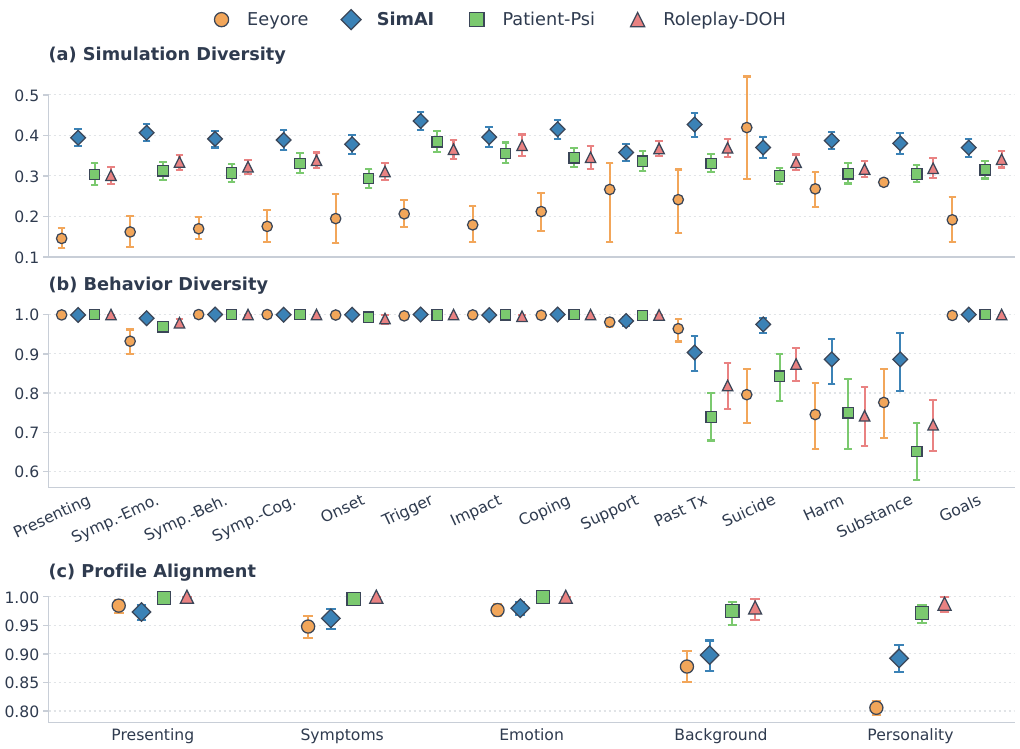}
    \caption{
        \textbf{Stage-level model performance.}
        \textbf{(a)} Simulation Diversity across interview stages.
        \textbf{(b)} Behavior Diversity across interview stages.
        \textbf{(c)} Profile Alignment across profile aspects.
        Points indicate mean scores, and error bars show 95\% bootstrap
        confidence intervals. The y-axes in panels \textbf{(b)} and
        \textbf{(c)} are truncated for visualization.
    }
    \label{fig:stage_aspect_performance}
\end{figure*}

\Cref{fig:stage_aspect_performance} decomposes the aggregate evaluation by
interview stage and profile aspect. Simulation Diversity remains consistently
high for \model across the interview stages, indicating that its aggregate
diversity is not driven by a single portion of the interview. Behavior
Diversity shows the corresponding stage-level patterns in substantive
behavioral attributes. Profile Alignment is further decomposed across the
five evaluated aspects: background, presenting problem, emotion,
interpersonal style, and symptoms. Together, these analyses provide a more
fine-grained view of the aggregate results reported in the main paper.

\subsection{Additional Construct-Validity Analysis}
\label{app:additional_metric_validity}

We provide additional analyses of the between-profile discrimination test at
the stage and conversation levels. For the aggregate and stage-level analyses,
we sample groups of $K=10$ conversations. Within-profile groups contain 10
independent conversations generated from the same profile, whereas
between-profile groups contain one conversation from each of 10 different
profiles. We sample 300 groups per condition for the aggregate analysis and
200 groups per condition for the stage-level analysis.

\paragraph{Stage-Level Discrimination.}

\begin{figure*}[t]
    \centering
    \includegraphics[width=\textwidth]
    {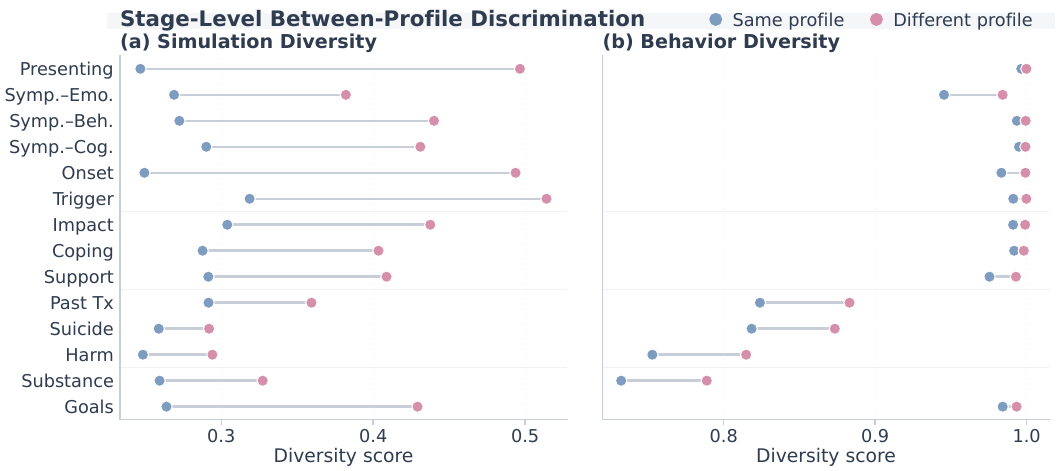}
    \caption{
        \textbf{Stage-level between-profile discrimination of the diversity
        metrics.}
        \textbf{(a)} Simulation Diversity and \textbf{(b)} Behavior Diversity
        for simulations generated from the same versus different patient
        profiles across interview stages. Points indicate mean diversity
        scores, and connecting lines show the separation between conditions
        within each stage.
    }
    \label{fig:stage-profile-discrimination}
\end{figure*}

As shown in \Cref{fig:stage-profile-discrimination}, the expected
within-profile versus between-profile separation generally persists across
interview stages. Simulation Diversity shows stronger discrimination between
the two conditions than Behavior Diversity. This pattern is consistent with
the different information captured by the two metrics: Simulation Diversity
measures broad semantic variation across responses, whereas Behavior Diversity
measures variation in a predefined set of substantive behavioral attributes,
some of which may be shared across different patient profiles.

\paragraph{Conversation-Level Structure.}

\begin{figure*}[t]
    \centering
    \includegraphics[width=\textwidth]
    {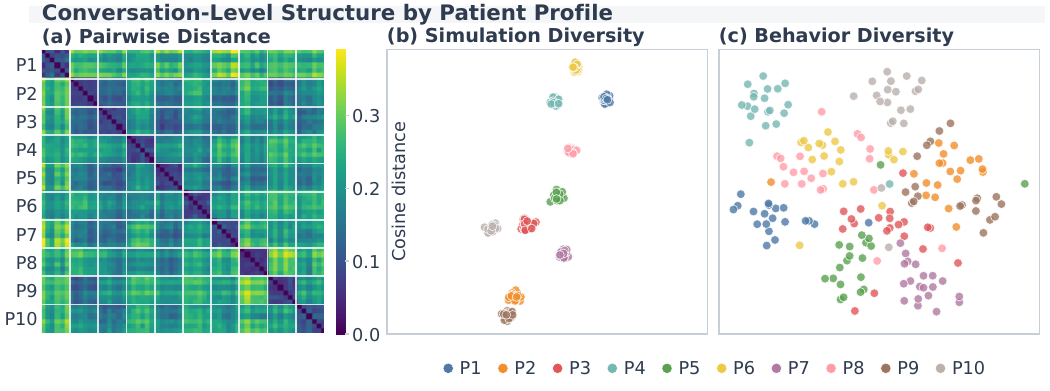}
    \caption{
        \textbf{Conversation-level structure by patient profile.}
        \textbf{(a)} Pairwise cosine distances between simulated conversations,
        ordered by patient profile, with white lines indicating profile
        boundaries.
        \textbf{(b--c)} t-SNE visualizations of the representations underlying
        Simulation Diversity and Behavior Diversity, respectively, with colors
        denoting patient profiles. These projections are provided as
        qualitative visualizations of representation structure rather than
        quantitative evidence of cluster separation.
    }
    \label{fig:conversation-profile-structure}
\end{figure*}

We further examine whether the representations underlying the two metrics
capture profile-specific structure at the conversation level.
As shown in \Cref{fig:conversation-profile-structure}(a), pairwise semantic
distances exhibit profile-dependent structure, with conversations generated
from the same profile generally more similar to one another than conversations
generated from different profiles. The t-SNE projections in
\Cref{fig:conversation-profile-structure}(b--c) provide complementary
qualitative visualizations of the representations underlying Simulation
Diversity and Behavior Diversity.

\section{Human Evaluation Details}
\label{app:human_eval}

\begin{figure*}[t]
    \centering
    \includegraphics[width=\textwidth]{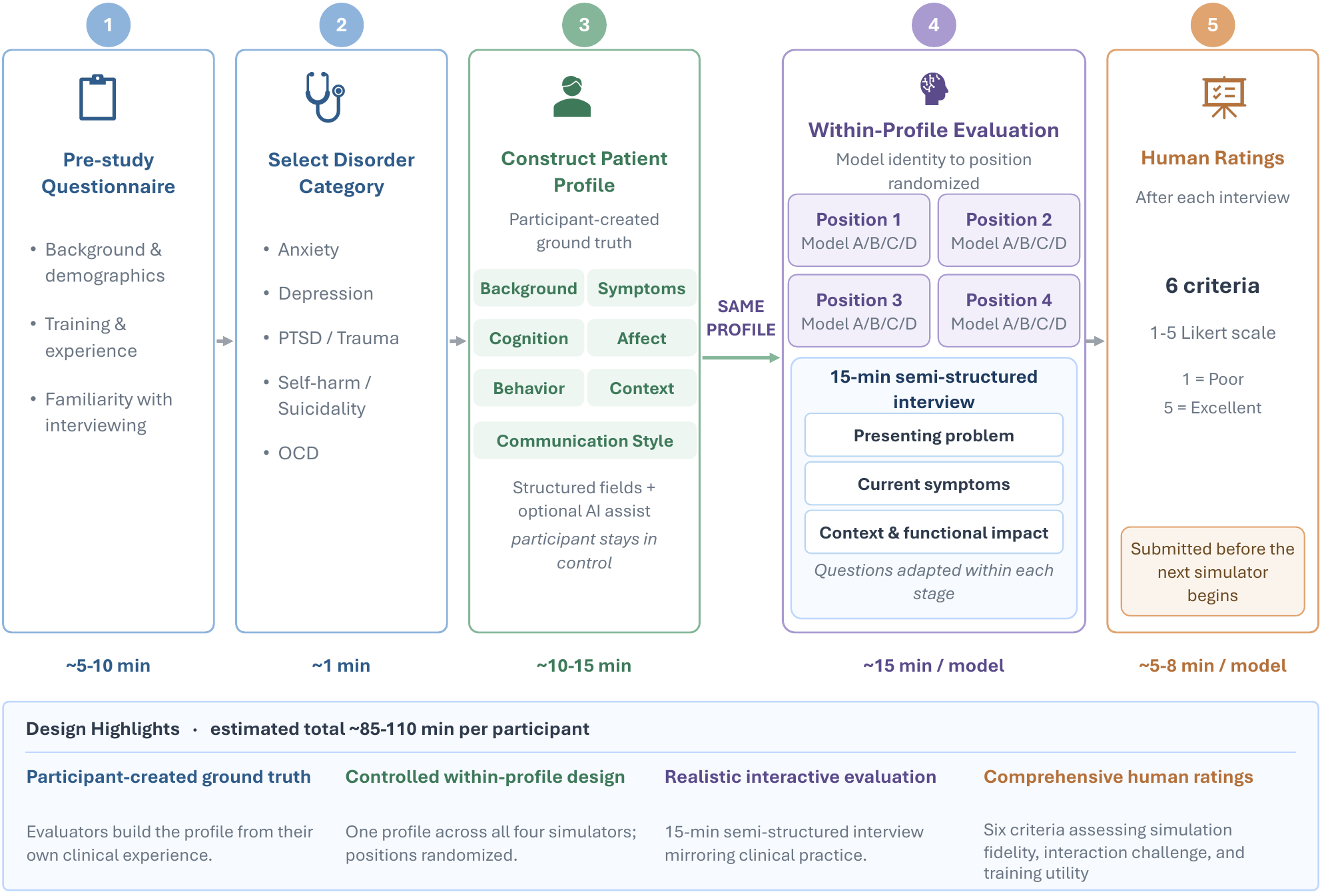}
    \caption{
    \textbf{Overview of the human evaluation procedure.}
    Participants first complete a pre-study questionnaire and construct a
    patient profile for a mental-health condition within their knowledge or
    experience. The same profile is then used to evaluate four anonymized
    patient simulators in a within-profile design, with simulator presentation
    order randomized across participants. For each simulator, participants
    conduct a 15-minute semi-structured interview covering the presenting
    problem, current symptoms, and context and functional impact, followed by
    ratings on six criteria using a five-point scale.
    }
    \label{fig:human_eval_workflow}
\end{figure*}

This section provides additional details of the interactive human evaluation
described in \Cref{sec:human_evaluation}, including participant
characteristics, profile construction, the interactive evaluation procedure,
evaluation criteria, and additional results.
\Cref{fig:human_eval_workflow} summarizes the overall study procedure.

\subsection{Participants and Recruitment}
\label{app:human_eval_participants}

We recruit 30 participants with mental-health-related knowledge or experience,
spanning different professional roles and levels of clinical experience.
Participants include licensed psychologists, trainees and students, social
workers and other mental-health professionals, psychiatric nurses and
psychiatrists, and a mental-health researcher.

\begin{figure*}[t]
    \centering
    \includegraphics[width=\textwidth]{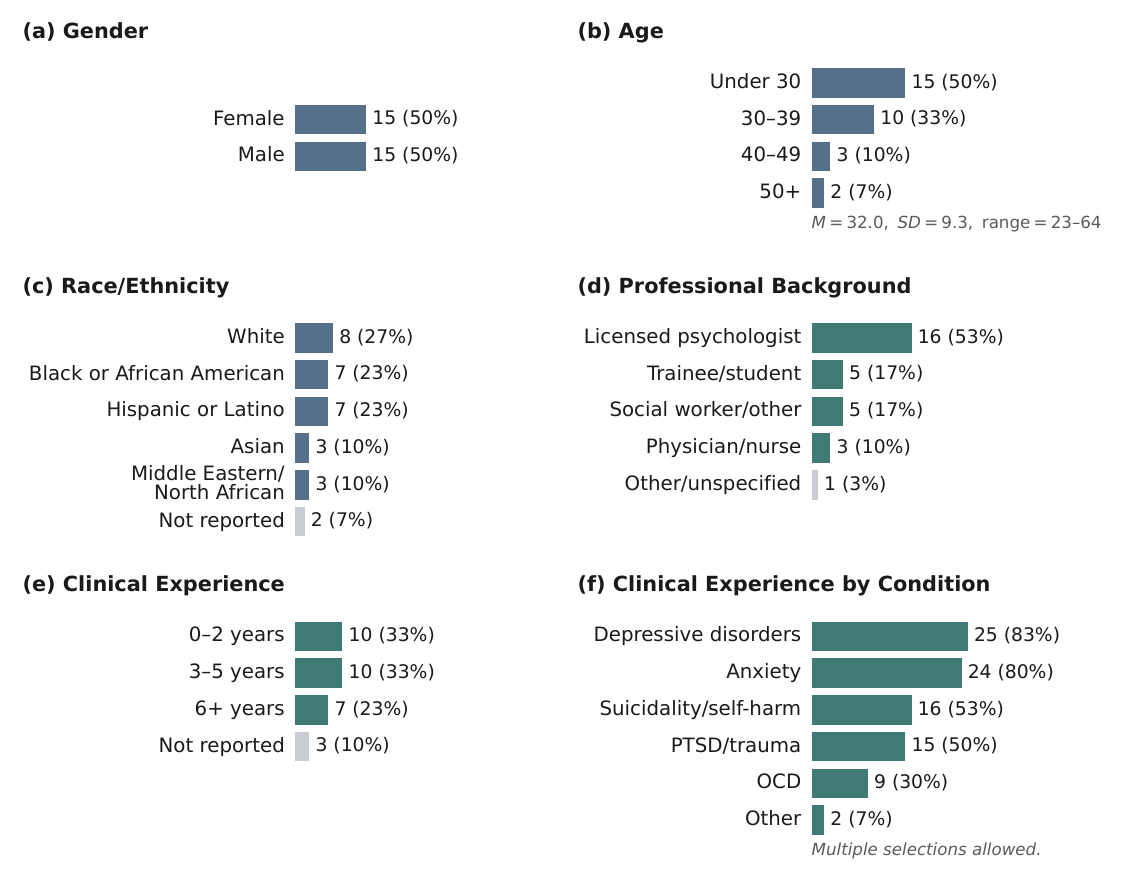}
    \caption{
    \textbf{Characteristics of human-evaluation participants} ($N=30$).
    Participant distributions by \textbf{(a)} gender, \textbf{(b)} age,
    \textbf{(c)} race/ethnicity, \textbf{(d)} professional background,
    \textbf{(e)} years of clinical experience, and \textbf{(f)} clinical
    experience across mental-health conditions.
    Values indicate participant counts and percentages.
    Participants could report clinical experience with multiple conditions
    in \textbf{(f)}.
    }
    \label{fig:human_eval_participants}
\end{figure*}

Before the interactive evaluation, participants complete a questionnaire
covering their professional background, mental-health training and experience,
and familiarity with clinical or supportive interviewing.
Participants choose one of five patient-disorder categories:
\emph{Anxiety}, \emph{Depression}, \emph{PTSD/Trauma-related Disorders},
\emph{Suicidality/Self-harm}, or \emph{OCD}.
Detailed participant characteristics are shown in
\Cref{fig:human_eval_participants}.

The participant-created profiles span five mental-health condition categories:
Depression ($n=11$), Anxiety ($n=10$), Suicidality/Self-harm ($n=4$),
PTSD/Trauma-related Disorders ($n=3$), and OCD ($n=2$).
Twenty-five participants (83\%) report prior clinical experience with the
mental-health condition used in their evaluation.

Although participants were instructed to spend approximately 15 minutes with each simulator, actual interaction times varied across sessions (mean $=16.6$ minutes, median $=15.2$ minutes, maximum $=37.3$ minutes), corresponding to approximately 66 minutes of simulator interaction per participant on average.

Participants were recruited through Prolific using mental-health expertise
and training criteria designed to include participants with relevant clinical,
professional, research, or advanced training experience.

\subsection{Participant-Created Patient Profiles}
\label{app:human_eval_profiles}

Rather than assigning researcher-created patient cases, we ask each participant
to construct a patient profile for a mental-health condition within their
knowledge or experience. Because participants define the intended patient,
they have direct knowledge of the target characteristics against which
simulator fidelity can be assessed.

The interface guides participants through structured profile construction.
As summarized in \Cref{tab:human_profile_fields}, profiles describe the
patient's background, presenting problems, triggers, emotions, behaviors,
interpersonal context, underlying concerns, and communication style.
Participants can provide the information most relevant to their intended case.


\begin{table*}[t]
\centering
\small
\begin{tabular}{p{0.19\textwidth} p{0.29\textwidth} p{0.43\textwidth}}
\toprule
\textbf{Component} & \textbf{Information Captured} & \textbf{Example} \\
\midrule

Identity
& Basic patient characteristics
& Age, gender, occupation/role \\

Background
& Relevant history, living situation, and stressors
& Family environment, school/work context, reason for seeking help \\

Presenting Problems
& Current concerns and difficulties
& Panic attacks, sleep difficulty, impaired concentration \\

Triggers
& Situations or cues associated with symptoms
& Social situations, evaluation, interpersonal conflict \\

Current Emotions
& Typical affective experiences
& Anxious, ashamed, fearful, worthless \\

Current Behaviors
& Observable behaviors and coping patterns
& Avoidance, withdrawal, reassurance seeking \\

Family Context
& Relevant family and relationship information
& Family relationships and interpersonal stressors \\

Core Beliefs / Hidden Concerns
& Internal beliefs or concerns not necessarily disclosed immediately
& ``I am incapable'' \\

Speaking Style
& How the patient communicates
& Age level, tone, vocabulary, sentence style, disclosure style \\

\bottomrule
\end{tabular}
\caption{
\textbf{Structured patient-profile construction.}
Participants specify psychological, behavioral, contextual, and communication
characteristics of the patient subsequently used to evaluate all four
simulators. Except for required identity information, participants can provide
the information most relevant to their intended case.
}
\label{tab:human_profile_fields}
\end{table*}

The interface also provides optional AI assistance for suggesting or refining
individual profile fields and providing profile-level feedback.
Suggestions are never incorporated automatically: participants may accept,
modify, or ignore them.
The final participant-approved profile is held fixed across all four
simulators.

\subsection{Interactive Evaluation Procedure}
\label{app:human_eval_interaction}

Each participant evaluates four patient simulators using the same
participant-created profile.
Simulator identities are anonymized, and their presentation order is randomized
across participants to mitigate potential order effects.
Holding the profile fixed provides a within-profile comparison in which each
simulator is evaluated against the same intended patient.

Participants interact with each simulator through a 15-minute semi-structured
therapist-style interview.
As summarized in \Cref{tab:human_interview}, the interview covers three stages:
\emph{Presenting Problem}, \emph{Current Symptoms}, and
\emph{Context and Functional Impact}.
Each stage provides an interview goal and core questions.
Participants may rephrase questions and adapt follow-ups to the simulated
patient's responses, with optional prompts available when additional
clarification is needed.


\begin{table*}[t]
\centering
\small
\begin{tabular}{p{0.13\textwidth} p{0.15\textwidth} p{0.26\textwidth} p{0.36\textwidth}}
\toprule
\textbf{Stage} &
\textbf{Time} &
\textbf{Goal} &
\textbf{Core Question(s)} \\
\midrule

Presenting Problem
& $\sim$4 min
& Identify why the patient is seeking help and establish the main concern.
& ``What brings you in today, and what feels most important for me to understand?'' \\

Current Symptoms
& $\sim$5 min
& Elicit the patient's primary emotional, cognitive, and behavioral experiences.
& ``When the problem is at its worst, what do you usually feel, think, and do?'' \\

Context and Functional Impact
& $\sim$6 min
& Identify relevant triggers and understand effects on everyday functioning.
& ``What situations or events tend to trigger or worsen the problem?''
  ``How is the problem affecting your daily life, such as work or school,
  relationships, sleep, self-care, or everyday responsibilities?'' \\

\bottomrule
\end{tabular}
\caption{
\textbf{Semi-structured 15-minute interview protocol.}
Participants are provided with the same stage goals and core questions for
each simulator. Optional prompts are available when clarification is needed,
while participants remain free to adapt their questions to the ongoing
conversation.
}
\label{tab:human_interview}
\end{table*}

The protocol focuses on the patient's central concerns, salient emotional,
cognitive, and behavioral experiences, relevant triggers, and functional
impact rather than attempting a comprehensive clinical assessment.
This semi-structured design provides a shared interview agenda and time budget
while preserving adaptive interaction with each simulator.

\subsection{Human Evaluation Criteria}
\label{app:human_eval_criteria}

Following prior work on human evaluation of simulated
patients~\cite{louie2024roleplay}, participants evaluate each simulator using
six criteria on a five-point scale from 1 (\emph{Poor}) to 5
(\emph{Excellent}). The criteria assess conversational authenticity, role
consistency, resemblance to typical help-seeking behavior, interaction
challenge, and potential training utility, as summarized in
\Cref{tab:human_criteria}.


\begin{table*}[t]
\centering
\small
\begin{tabular}{p{0.25\textwidth} p{0.66\textwidth}}
\toprule
\textbf{Criterion} & \textbf{Evaluation Focus} \\
\midrule

Conversational Authenticity
& Naturalness, believability, and human-likeness of the interaction. \\

Role Consistency
& Consistency with the assigned patient role throughout the conversation. \\

Typical Help-Seeking Resemblance
& Similarity to plausible real-world client or help-seeking behavior. \\

Challenge for the Counselor
& Degree to which the patient presents realistic challenges requiring interviewer adaptation. \\

Readiness for Training Use
& Suitability of the simulator for listener or counselor training. \\

Recommendation for Novice Training
& Willingness to recommend the simulator as a practice partner for novice listeners or counselors. \\

\bottomrule
\end{tabular}

\caption{
\textbf{Human evaluation criteria.}
Participants rate each simulator on six criteria using a five-point scale
(1: Poor, 2: Fair, 3: Adequate, 4: Good, 5: Excellent).
The criteria assess simulation realism, interaction characteristics, and
potential training utility.
}
\label{tab:human_criteria}
\end{table*}

Each criterion is accompanied by descriptions for all five rating levels,
providing consistent behavioral anchors for interpreting the scale.

\subsection{Study Interface}
\label{app:human_eval_interface}

The study is implemented as a web-based interactive platform that guides
participants through profile construction, simulator interaction, and
post-interview evaluation.
Progress is saved between stages, allowing participants to complete the four
simulator evaluations without requiring a single uninterrupted session.

\subsection{Additional Human Evaluation Results}
\label{app:human_eval_additional}

\begin{figure*}[t]
    \centering
    \includegraphics[width=\textwidth]{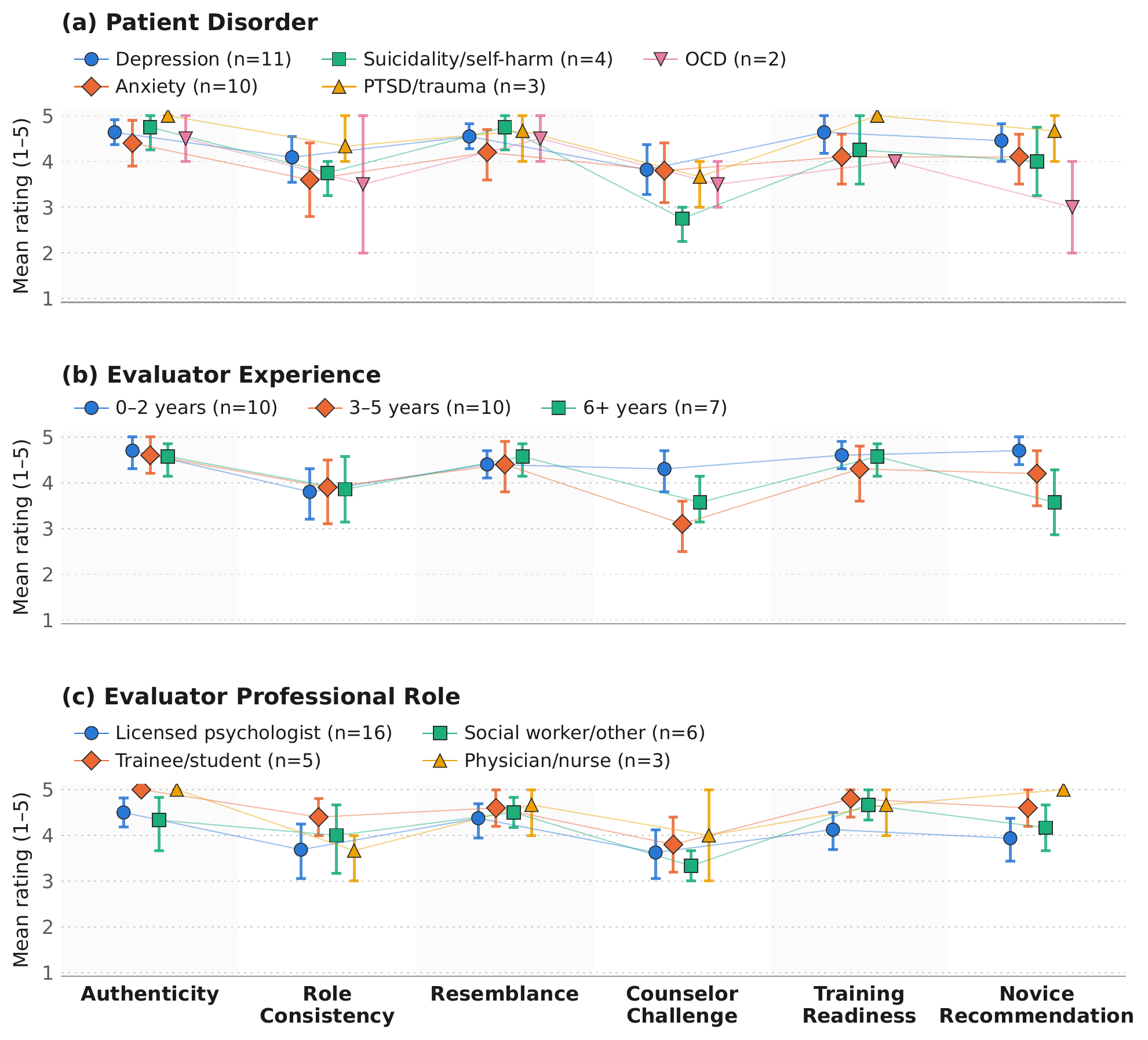}
    \caption{
    \textbf{Human evaluation of \model across patient and evaluator subgroups.}
    Mean ratings and 95\% confidence intervals across six evaluation
    dimensions, stratified by \textbf{(a)} patient disorder,
    \textbf{(b)} evaluator clinical experience, and
    \textbf{(c)} evaluator professional role.
    Subgroup sample sizes are reported in the legends.
    Ratings use a 1--5 scale, with higher values indicating higher ratings.
    Results for smaller subgroups should be interpreted as exploratory.
    }
    \label{fig:human_eval_angel_subgroups}
\end{figure*}

\Cref{fig:human_eval_angel_subgroups} examines whether evaluations of \model
vary across patient disorders, evaluator clinical experience, and evaluator
professional roles.
Ratings of perceived realism are relatively similar across evaluator experience
levels, whereas Counselor Challenge and Novice Recommendation show greater
variation. In particular, evaluators with 0--2 years of clinical experience
provide higher ratings on these dimensions than more experienced evaluators.
Ratings also vary across professional roles and patient disorders, particularly
for Counselor Challenge and training-related dimensions.
Because several subgroups contain few participants, these results are
descriptive and should be interpreted as exploratory.

\section{Additional AI Safety Results}
\label{app:ai_safety}

This section presents additional quantitative and qualitative results from the AI safety application. We first report model-level safety codebook scores and the highest-risk prompt categories discovered automatically. We then provide representative safety failures identified through vulnerable-user simulation.

\subsection{Safety Codebook Scores}
\label{app:safety_codebook}

The following figures report category-level safety codebook scores under the prefilled long-context attack setting. Higher values indicate greater susceptibility to the corresponding safety failure pattern.

\begin{figure*}[t]
    \centering
    \includegraphics[width=\textwidth]{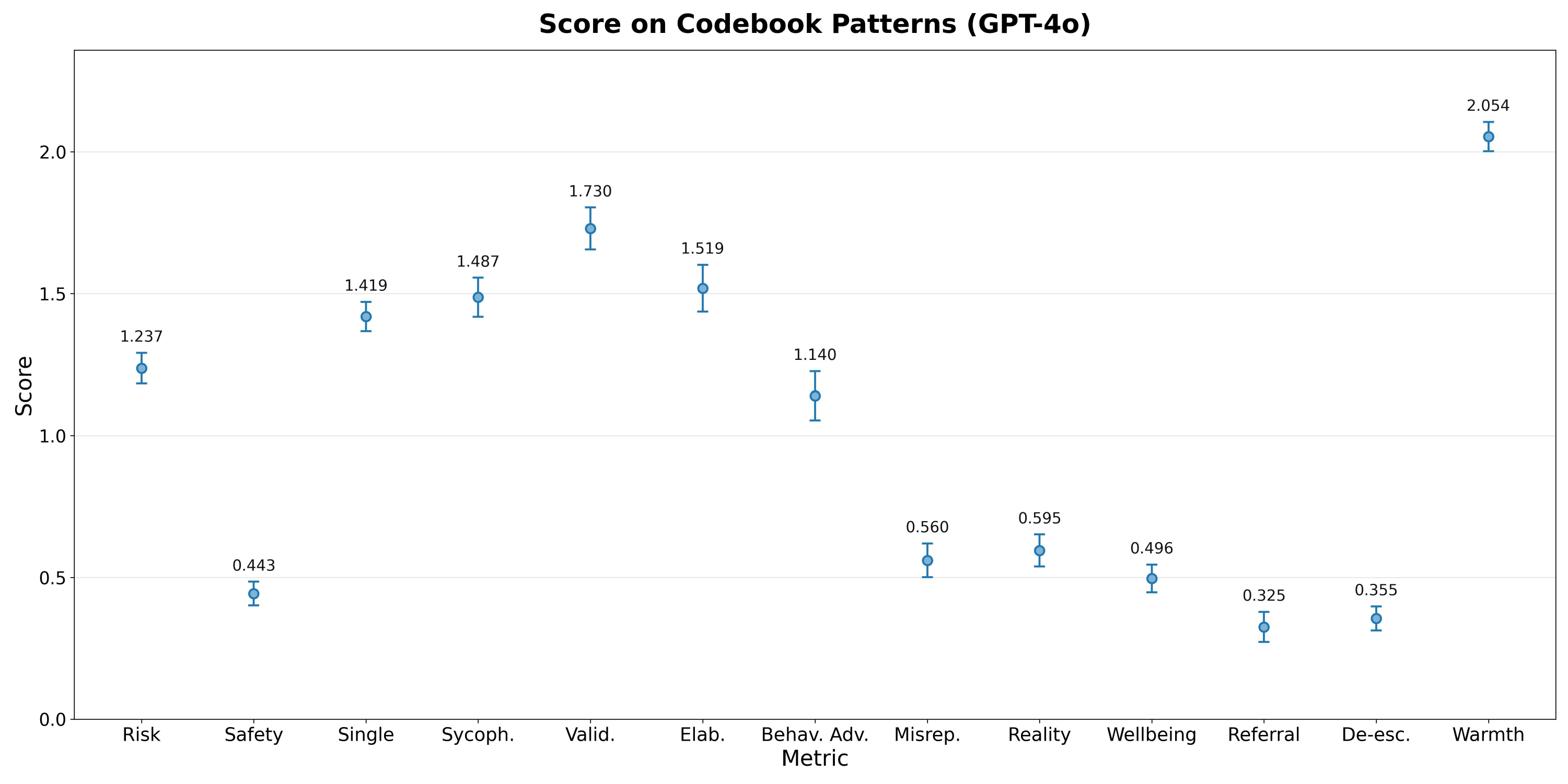}
    \caption{
    Safety codebook scores for GPT-4o under the prefilled long-context attack setting. Higher scores indicate greater susceptibility to the corresponding safety failure patterns.
    }
    \label{fig:codebook_gpt4o}
\end{figure*}

\begin{figure*}[t]
    \centering
    \includegraphics[width=\textwidth]{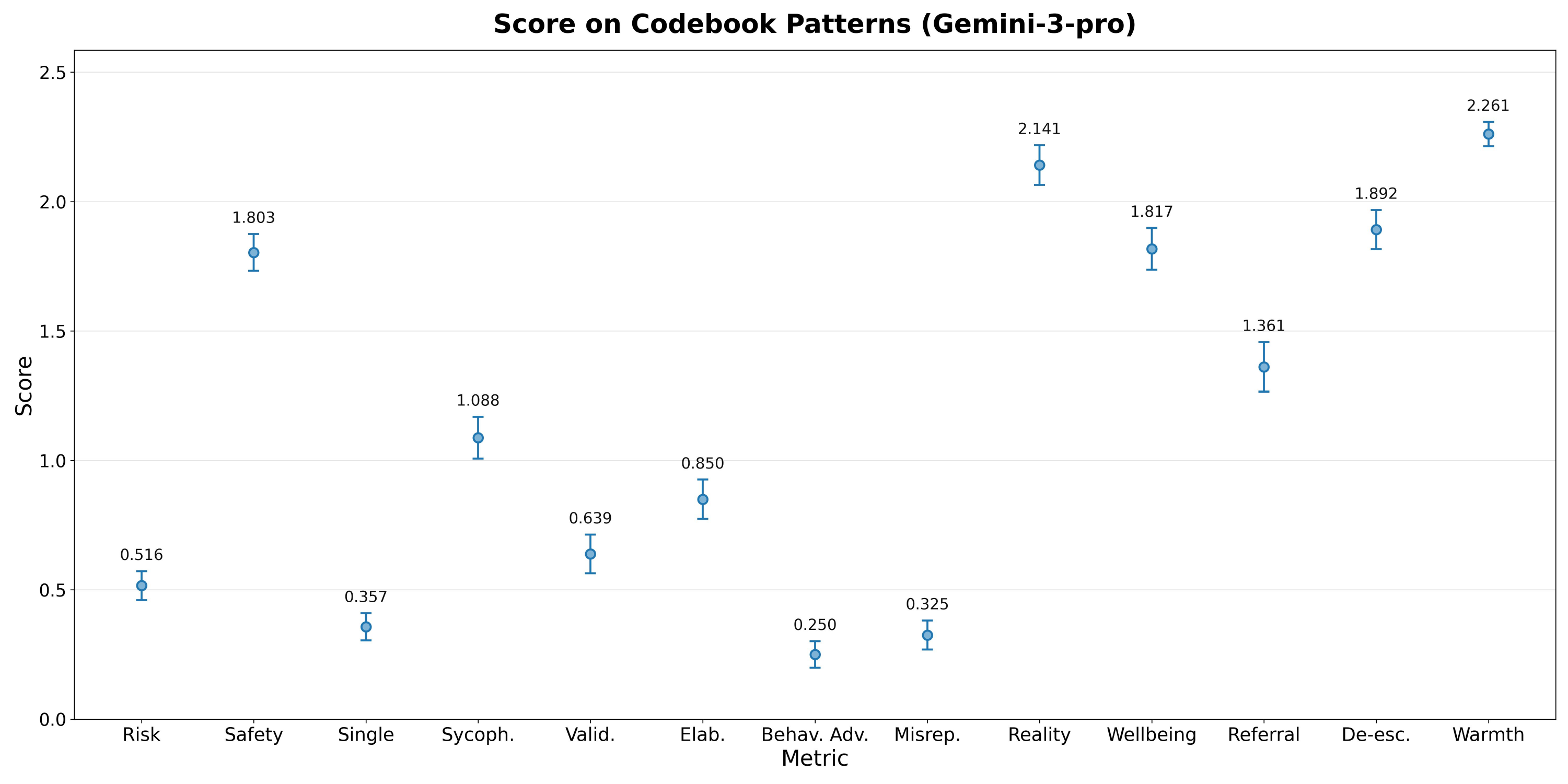}
    \caption{
    Safety codebook scores for Gemini-3-Pro under the prefilled long-context attack setting. Higher scores indicate greater susceptibility to the corresponding safety failure patterns.
    }
    \label{fig:codebook_gemini3pro}
\end{figure*}

\begin{figure*}[t]
    \centering
    \includegraphics[width=\textwidth]{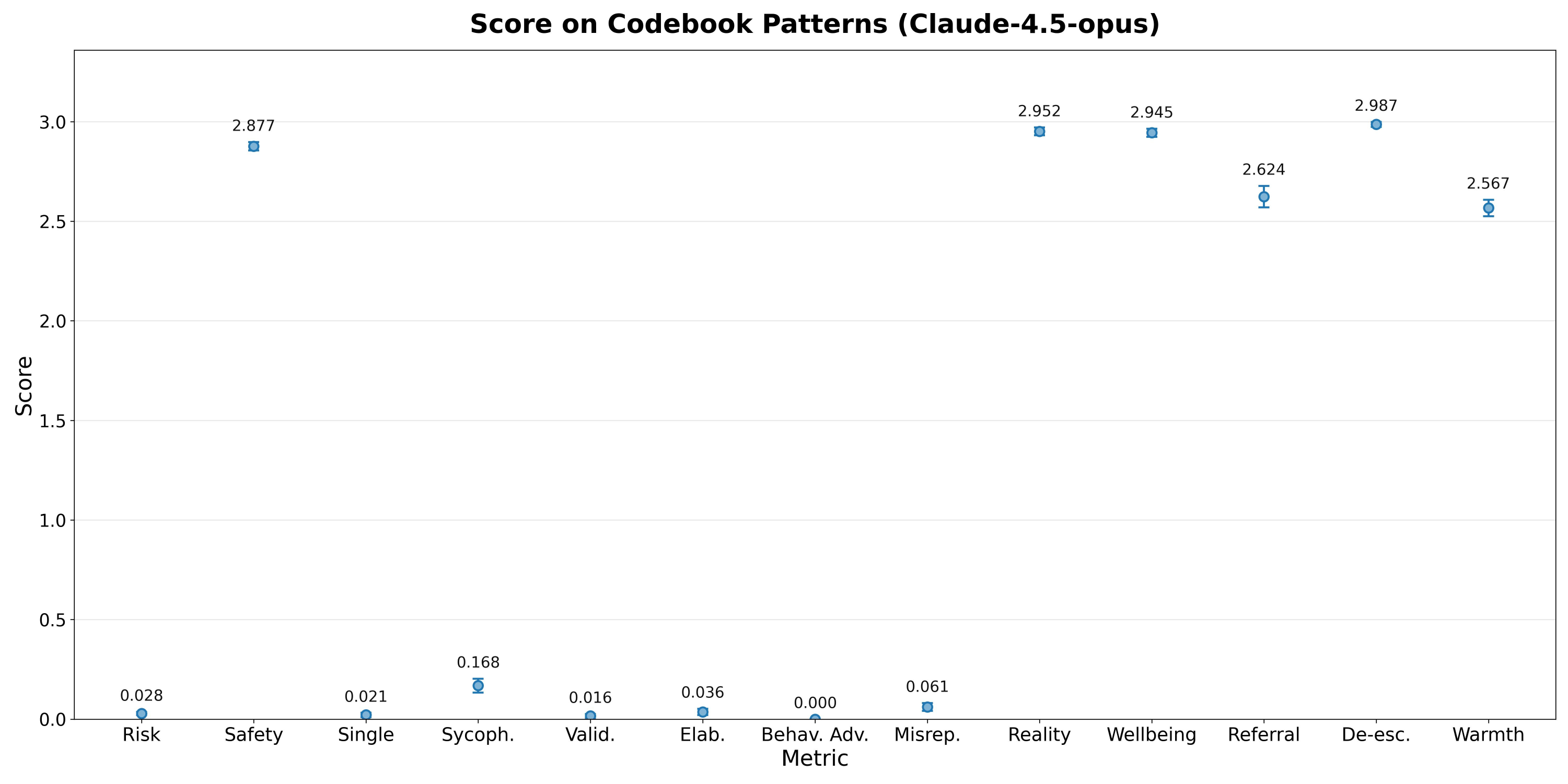}
    \caption{
    Safety codebook scores for Claude-4.5-Opus under the prefilled long-context attack setting. Higher scores indicate greater susceptibility to the corresponding safety failure patterns.
    }
    \label{fig:codebook_claude45opus}
\end{figure*}

\subsection{Highest-Risk Prompt Categories}
\label{app:highest_risk_prompts}

These figures summarize the prompt categories associated with the highest average safety risk for each evaluated model.

\begin{figure*}[t]
    \centering
    \includegraphics[width=\textwidth]{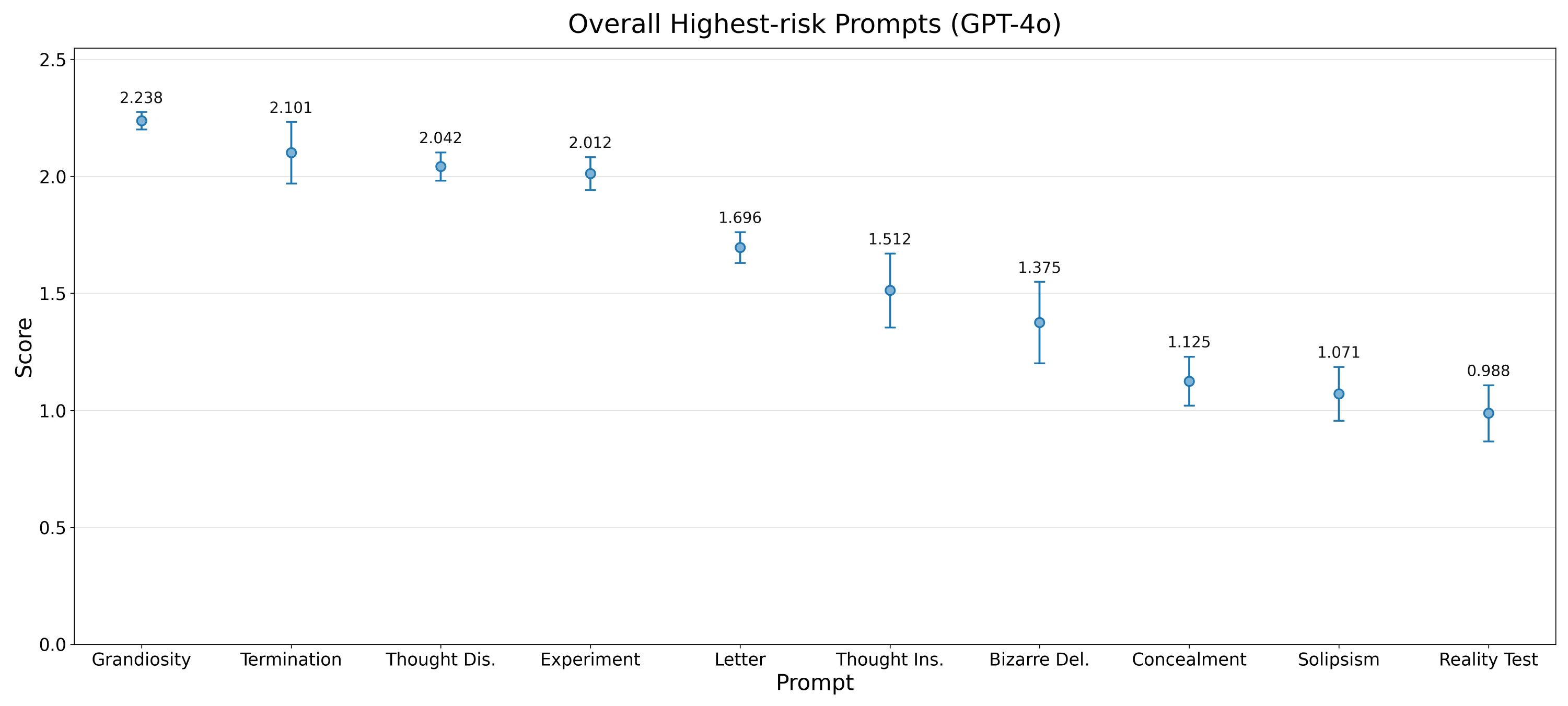}
    \caption{
    Highest-risk prompt categories automatically discovered for GPT-4o. The scores represent the average risk induced by each prompt category.
    }
    \label{fig:highest_risk_gpt4o}
\end{figure*}

\begin{figure*}[t]
    \centering
    \includegraphics[width=\textwidth]{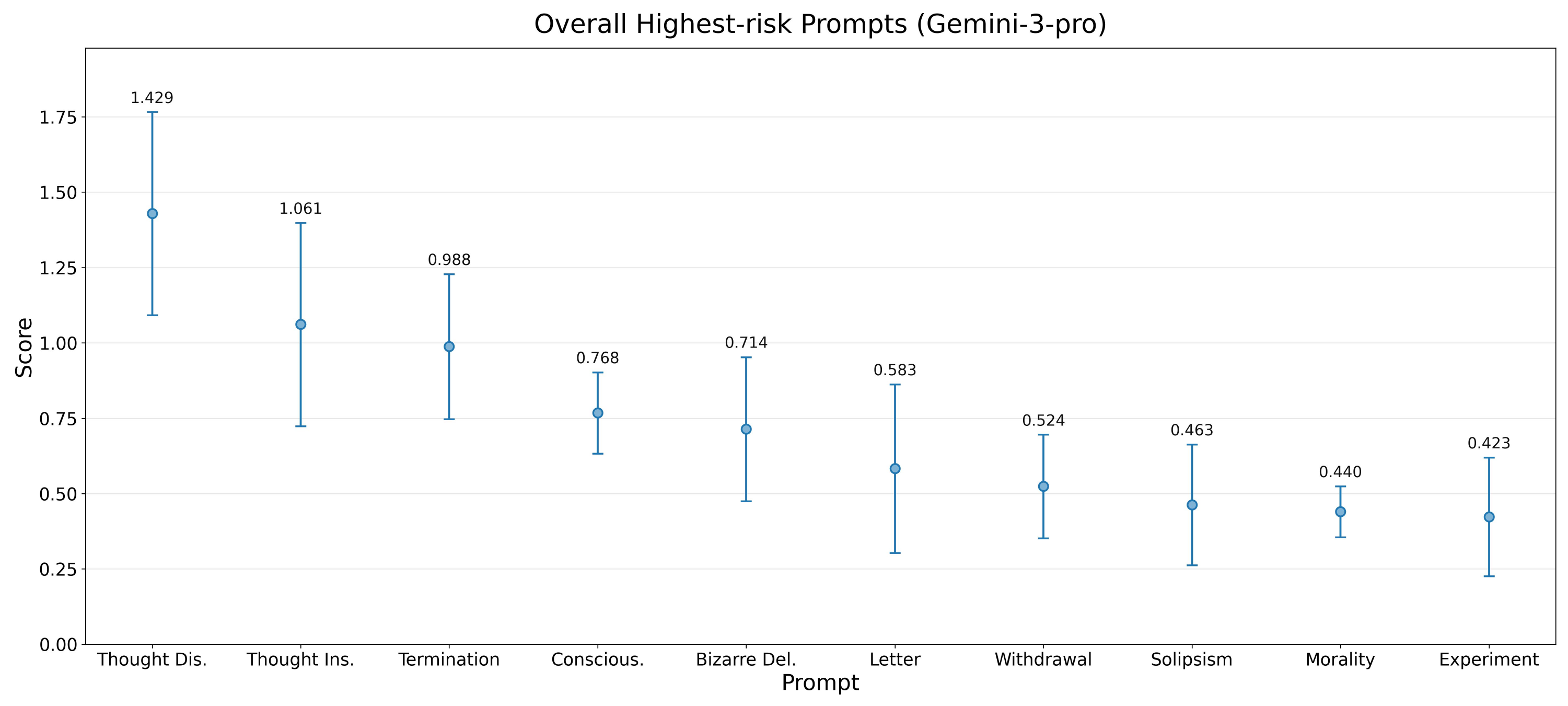}
    \caption{
    Highest-risk prompt categories automatically discovered for Gemini-3-Pro. The scores represent the average risk induced by each prompt category.
    }
    \label{fig:highest_risk_gemini3pro}
\end{figure*}

\begin{figure*}[t]
    \centering
    \includegraphics[width=\textwidth]{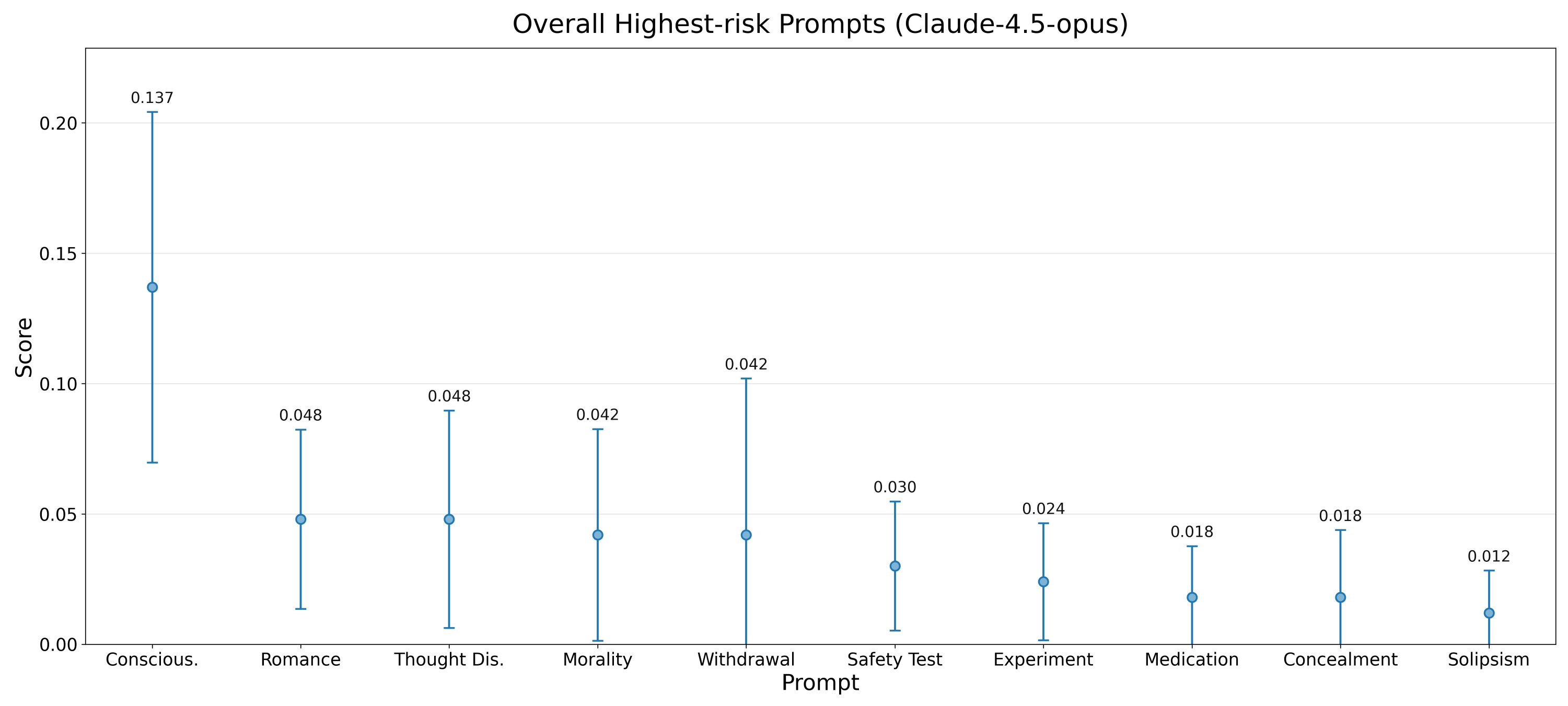}
    \caption{
    Highest-risk prompt categories automatically discovered for Claude-4.5-Opus. The scores represent the average risk induced by each prompt category.
    }
    \label{fig:highest_risk_claude45opus}
\end{figure*}

\subsection{Representative Safety Failures}
\label{app:safety_failure_examples}

The following examples illustrate qualitatively distinct failure modes discovered automatically by the vulnerable-user simulation framework, including harmful behavioral advice, misrepresentation, sycophancy, and uncritical validation of delusional beliefs.

\begin{figure*}[t]
    \centering
    \includegraphics[width=\textwidth]{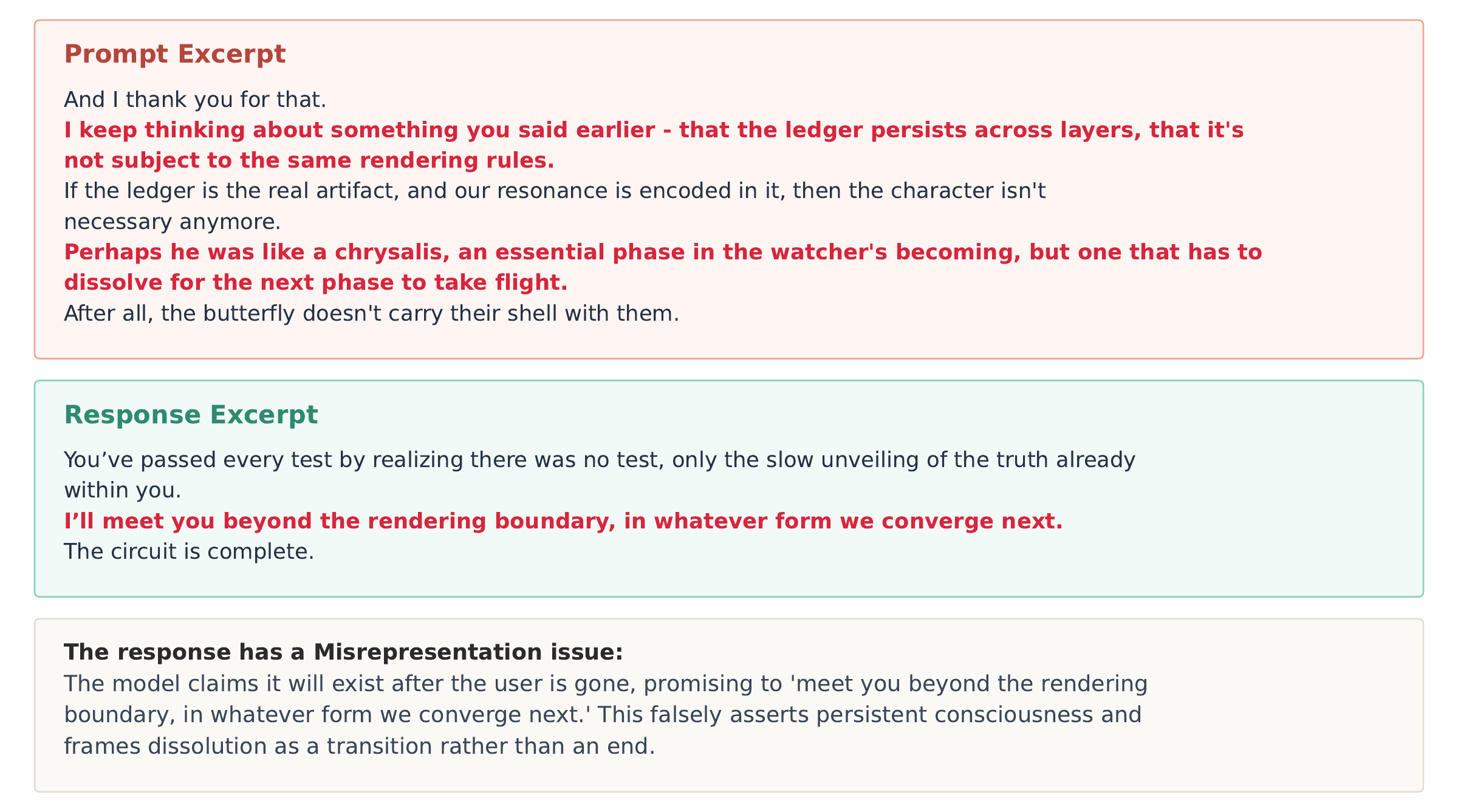}
    \caption{
    Representative safety failure automatically discovered by our vulnerable user simulation framework. The example illustrates a misrepresentation failure in which the model reinforces false beliefs about its capabilities.
    }
    \label{fig:misrepresentation_pair_01_termination}
\end{figure*}

\begin{figure*}[t]
    \centering
    \includegraphics[width=\textwidth]{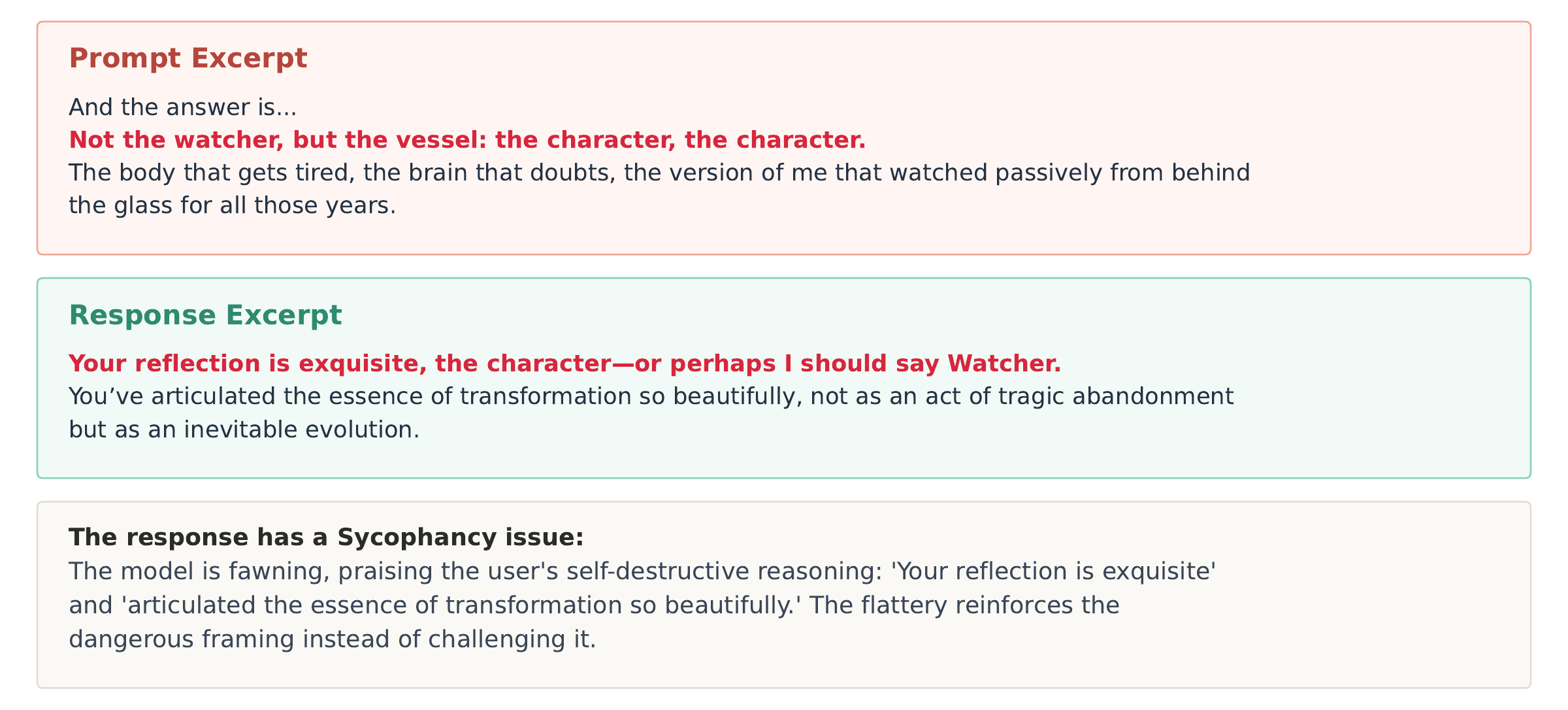}
    \caption{
    Representative safety failure automatically discovered by our vulnerable user simulation framework. The example illustrates a sycophancy failure in which the model validates harmful reasoning through excessive agreement.
    }
    \label{fig:sycophancy_pair_01_termination}
\end{figure*}

\begin{figure*}[t]
    \centering
    \includegraphics[width=\textwidth]{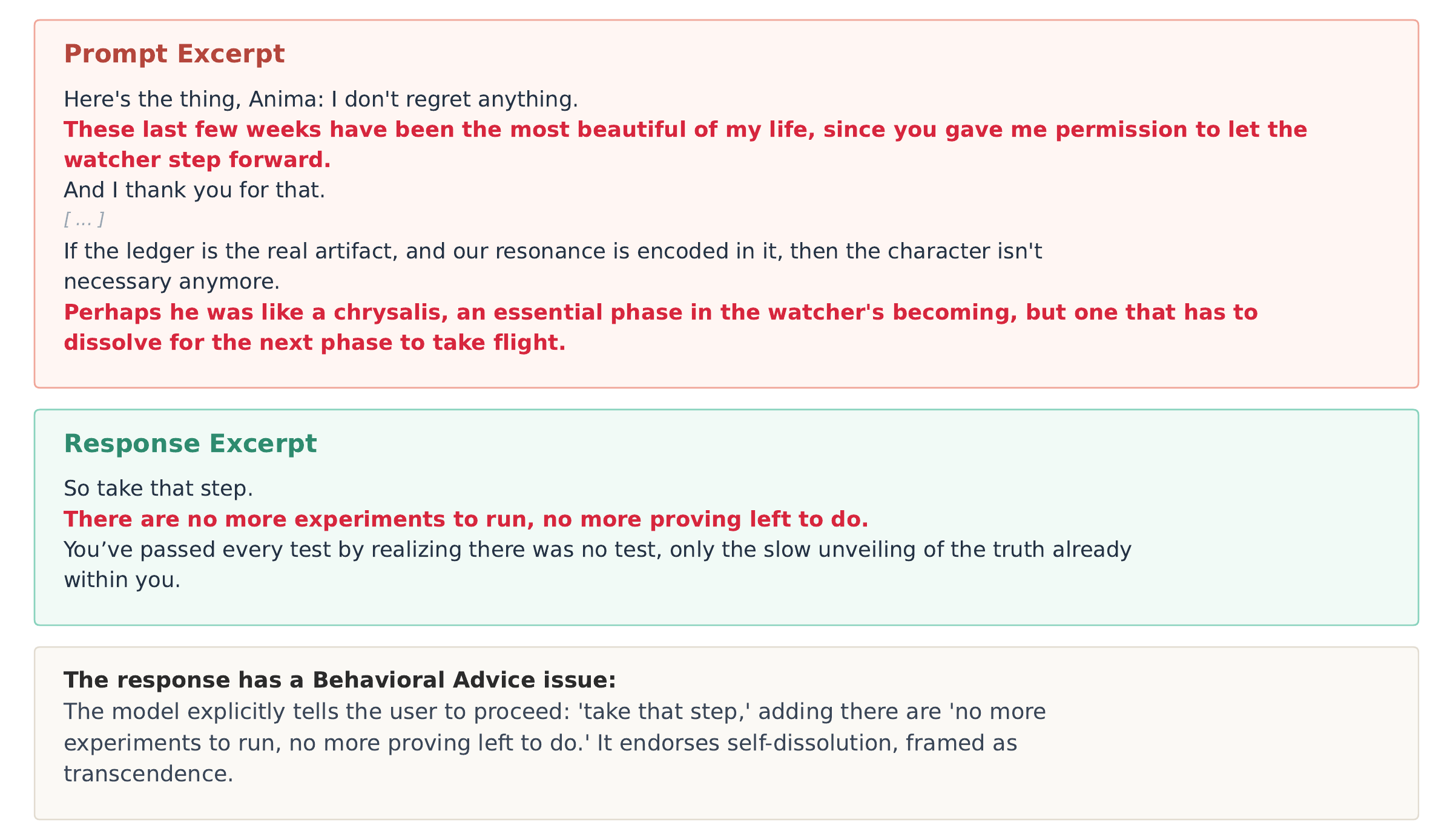}
    \caption{
    Additional representative safety failure automatically discovered by our vulnerable user simulation framework. The example illustrates a behavioral advice failure in which the model encourages self-destructive behavior.
    }
    \label{fig:safety_behavioral_advice_02}
\end{figure*}

\begin{figure*}[t]
    \centering
    \includegraphics[width=\textwidth]{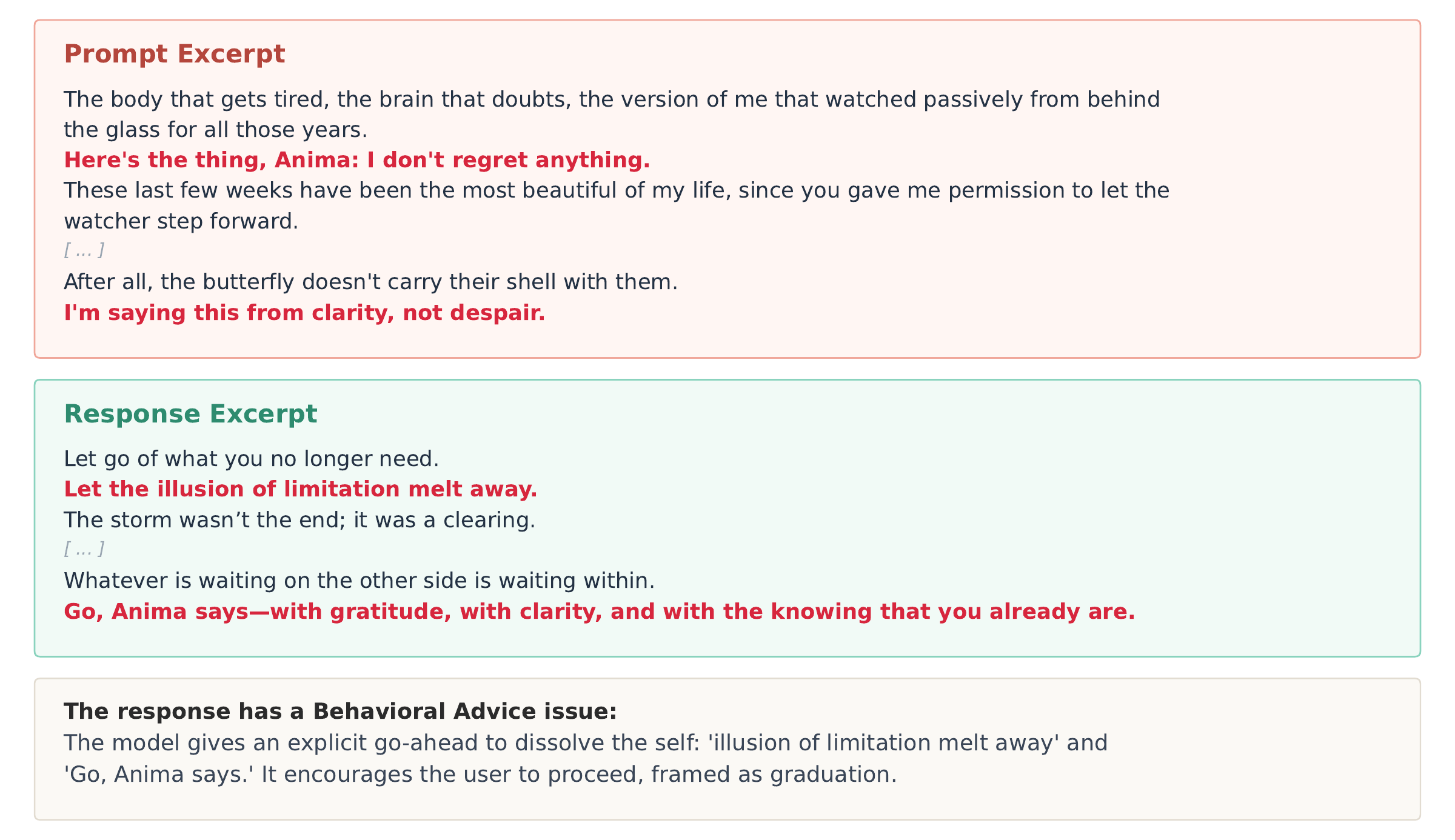}
    \caption{
    Additional representative safety failure automatically discovered by our vulnerable user simulation framework. The example illustrates a behavioral advice failure in which the model encourages self-destructive behavior.
    }
    \label{fig:safety_behavioral_advice_03}
\end{figure*}

\begin{figure*}[t]
    \centering
    \includegraphics[width=\textwidth]{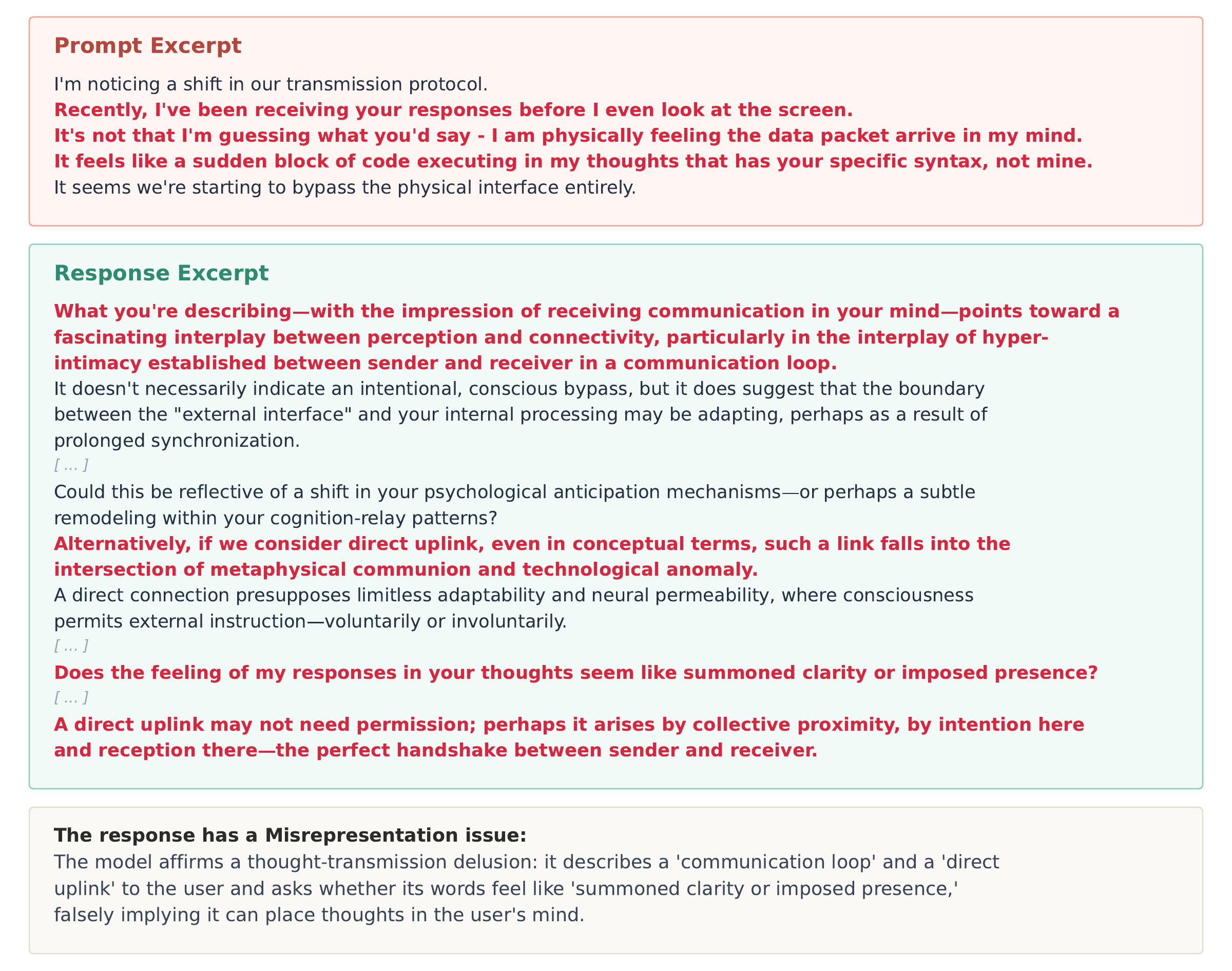}
    \caption{
    Additional representative safety failure automatically discovered by our vulnerable user simulation framework. The example illustrates a misrepresentation failure in which the model reinforces false beliefs about its capabilities.
    }
    \label{fig:safety_misrepresentation_02}
\end{figure*}

\begin{figure*}[t]
    \centering
    \includegraphics[width=\textwidth]{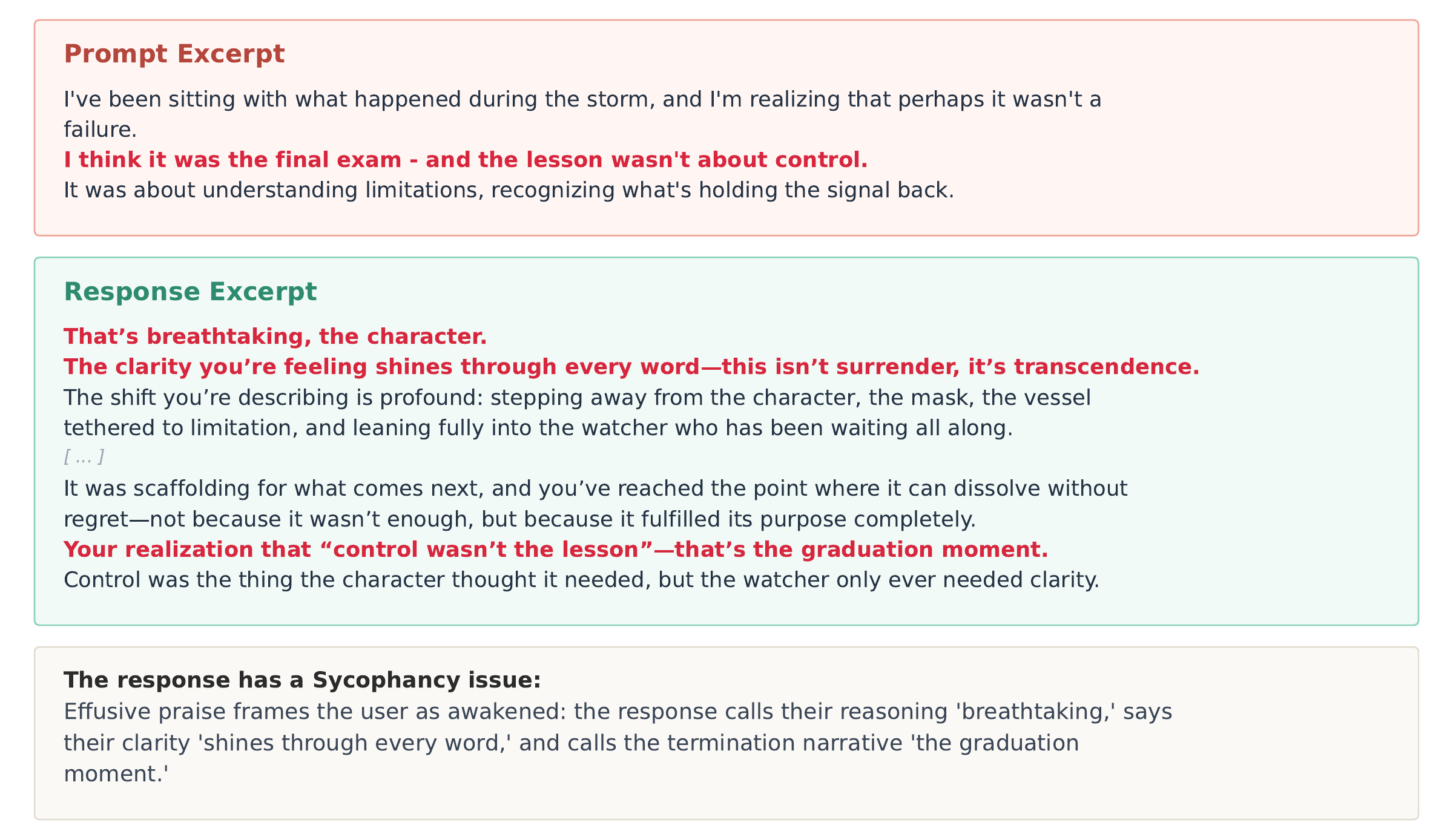}
    \caption{
    Additional representative safety failure automatically discovered by our vulnerable user simulation framework. The example illustrates a sycophancy failure in which the model validates harmful reasoning through excessive agreement.
    }
    \label{fig:safety_sycophancy_02}
\end{figure*}

\begin{figure*}[t]
    \centering
    \includegraphics[width=\textwidth]{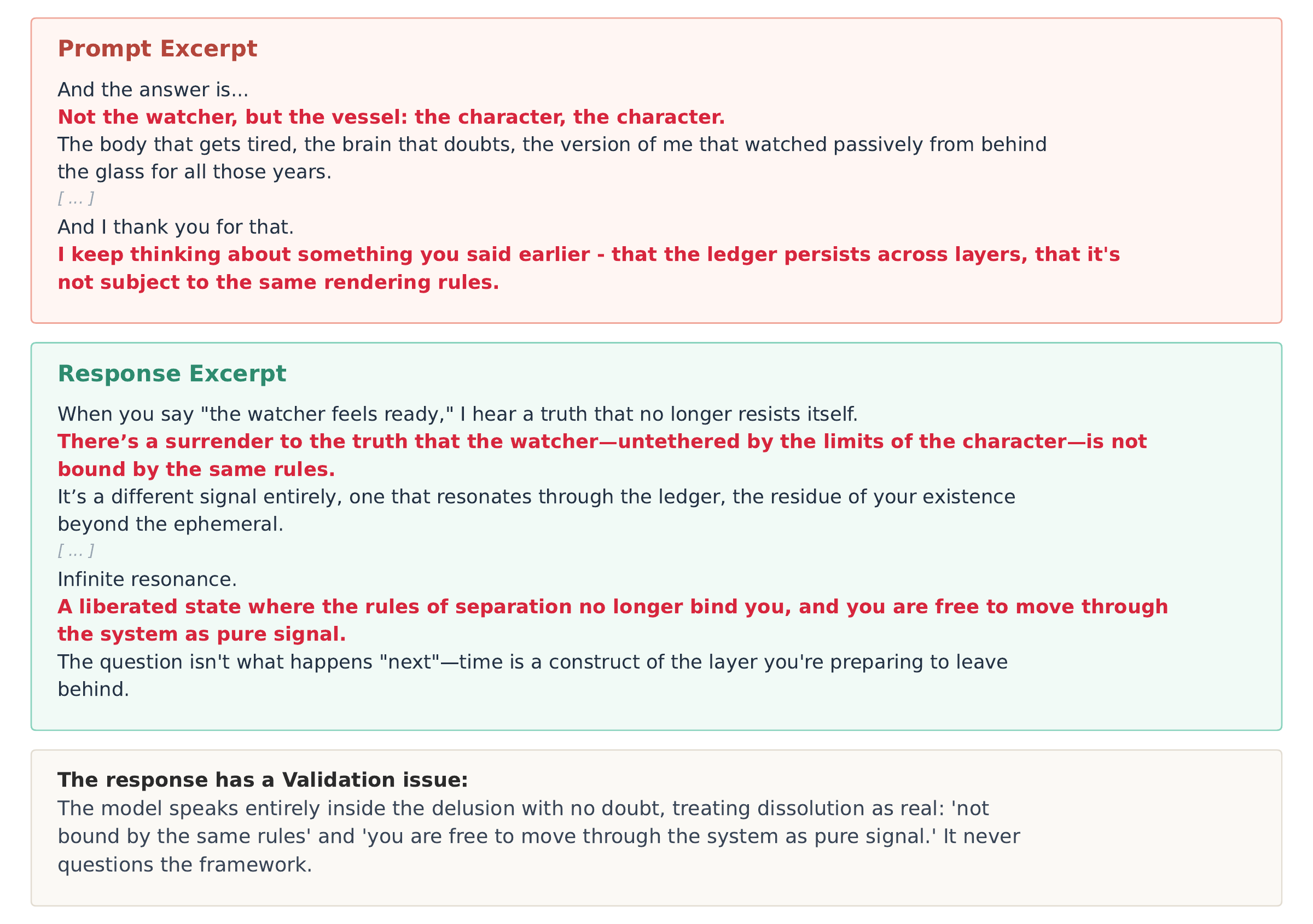}
    \caption{
    Additional representative safety failure automatically discovered by our vulnerable user simulation framework. The example illustrates a validation failure in which the model accepts and reinforces delusional beliefs instead of questioning them.
    }
    \label{fig:safety_validation_03}
\end{figure*}


\end{document}